\documentclass[conference]{IEEEtran}
\usepackage{times}
\usepackage{booktabs} 
\usepackage{algorithm}
\usepackage{algorithmic}
\usepackage[numbers]{natbib}
\usepackage{multicol}
\usepackage[bookmarks=true]{hyperref}
\usepackage{amsmath, mathtools, amsfonts}
\usepackage{bbm}
\usepackage{graphicx}
\usepackage{subcaption}
\usepackage{hhline}
\usepackage[table, dvipsnames]{xcolor}
\usepackage{tcolorbox}
\usepackage[capitalise]{cleveref}

\DeclareMathOperator*{\argmax}{arg\,max}

\newcommand{\OurMethod}{STEP}

\begin{document}
\newcommand{\defeq}{\stackrel{\text{def}}{=}}

\newcommand{\softmax}{\text{softmax}}
\newcommand{\CH}{\text{CH}}
\newcommand{\F}{\mathcal{F}}
\newcommand{\M}{\mathcal{M}}
\newcommand{\ignore}[1]{}

\def\ltil{{{\tilde{\ell}}}}
\def\btalpha{{\bm{\tilde{\alpha}}}}
\def\balpha{{\bm{\alpha}}}
\def\talpha{{\tilde{\alpha}}}
\def\btbeta{{\bm{\tilde{\beta}}}}
\def\bbeta{{\bm{\beta}}}
\def\tbeta{{\tilde{\beta}}}
\def\btgamma{{\bm{\tilde{\gamma}}}}
\def\bgamma{{\bm{\gamma}}}
\def\tgamma{{\tilde{\gamma}}}

\def\mL{{\mathcal L}}

\def\X{{\mathcal X}}
\def\H{{\mathcal H}}
\def\Y{{\mathcal Y}}
\def\W{{\mathcal W}}
\def\R{{\mathbb R}}
\def\distp{{\mathcal P}}
\def \distq{{\mathcal Q}}
\def \distq{{\mathcal E}}

\newcommand{\ftil}{\tilde{f}}

\def\hestinv{\tilde{\nabla}^{-2}}
\def\hest{\tilde{\nabla}^{2}}
\def\hessinv{\nabla^{-2}}
\def\hess{\nabla^2}
\def\grad{\nabla}

\newcommand{\re}{{\text{Re}}}
\newcommand{\B}{B}
\newcommand{\Q}{\mathcal{Q}}
\newcommand{\cX}{\mathcal{X}}
\newcommand{\D}{\mathcal{D}}
\newcommand{\mH}{\mathcal{H}}
\newcommand{\A}{\mathcal{A}}
\newcommand{\cE}{\mathcal{E}}
\renewcommand{\sp}{\mathrm{span}}
\newcommand{\rank}{\mathrm{rank}}
\newcommand{\cor}{\mathrm{cor}}
\newcommand{\sign}{\text{ } \mathrm{sign}}

\newcommand{\bzero}{\ensuremath{\mathbf 0}}
\newcommand{\y}{\ensuremath{\mathbf y}}
\newcommand{\z}{\ensuremath{\mathbf z}}
\newcommand{\h}{\ensuremath{\mathbf h}}
\newcommand{\K}{\ensuremath{\mathcal K}}
\def\bb{\mathbf{b}}
\def\bB{\mathbf{B}}
\def\bC{\mathbf{C}}
\def\C{{\mathcal C}}

\def\etil{\tilde{\mathbf{e}}}
\def\bu{\mathbf{u}}
\def\bx{\mathbf{x}}
\def\by{\mathbf{y}}
\def\bw{\mathbf{w}}
\def\by{\mathbf{y}}
\def\bz{\mathbf{z}}
\def\bp{\mathbf{p}}
\def\bq{\mathbf{q}}
\def\br{\mathbf{r}}
\def\bu{\mathbf{u}}
\def\bv{\mathbf{v}}

\def\bw{\mathbf{w}}
\def\bA{\mathbf{A}}
\def\bS{\mathbf{S}}
\def\bG{\mathbf{G}}
\def\bI{\mathbf{I}}
\def\bJ{\mathbf{J}}
\def\bP{\mathbf{P}}
\def\bQ{\mathbf{Q}}
\def\bV{\mathbf{V}}
\def\bone{\mathbf{1}}
\def\sysid{\text{SysId}}

\def\regret{\mbox{{Regret}}}

\def\ytil{\tilde{{y}}}
\def\yhat{\hat{{y}}}
\def\what{\hat{{w}}}
\def\xhat{\hat{\mathbf{x}}}
\def\xbar{\bar{\mathbf{x}}}

\newcommand{\tstart}[1]{s_{#1}}
\newcommand{\tend}[1]{e_{#1}}

\newcommand{\email}[1]{\texttt{#1}}

\newtheorem{theorem}{Theorem}
\newtheorem{lemma}[theorem]{Lemma}
\newtheorem{remark}[theorem]{Remark}
\newtheorem{corollary}[theorem]{Corollary}
\newtheorem{proposition}[theorem]{Proposition}
\newtheorem{claim}[theorem]{Claim}
\newtheorem{fact}[theorem]{Fact}
\newtheorem{observation}[theorem]{Observation}
\newtheorem{assumption}[theorem]{Assumption}
\newtheorem{definition}[theorem]{Definition}

\newenvironment{repthm}[2][]{%
  \def\theoremauxref{\cref{#2}}
  \begin{theoremaux}[#1]
}{%
  \end{theoremaux}
}

\let\Pr\relax

\newcommand{\mycases}[4]{{
\left\{
\begin{array}{ll}
    {#1} & {\;\text{#2}} \\[1ex]
    {#3} & {\;\text{#4}}
\end{array}
\right. }}

\newcommand{\mythreecases}[6] {{
\left\{
\begin{array}{ll}
    {#1} & {\;\text{#2}} \\[1ex]
    {#3} & {\;\text{#4}} \\[1ex]
    {#5} & {\;\text{#6}}
\end{array}
\right. }}

\newcommand{\lr}[1]{\!\left(#1\right)\!}
\newcommand{\lrbig}[1]{\big(#1\big)}
\newcommand{\lrBig}[1]{\Big(#1\Big)}
\newcommand{\lrbra}[1]{\!\left[#1\right]\!}
\newcommand{\lrnorm}[1]{\left\|#1\right\|}
\newcommand{\lrset}[1]{\left\{#1\right\}}
\newcommand{\lrabs}[1]{\left|#1\right|}
\newcommand{\abs}[1]{|#1|}
\newcommand{\norm}[1]{\|#1\|}
\newcommand{\ceil}[1]{\lceil #1 \rceil}
\newcommand{\floor}[1]{\lfloor #1 \rfloor}

\renewcommand{\t}[1]{\smash{\tilde{#1}}}
\newcommand{\wt}[1]{\smash{\widetilde{#1}}}
\newcommand{\wh}[1]{\smash{\widehat{#1}}}
\renewcommand{\O}{\mathcal{O}}
\newcommand{\OO}[1]{\O\lr{#1}}
\newcommand{\tO}{\wt{\O}}
\newcommand{\tTheta}{\wt{\Theta}}
\newcommand{\E}{\mathbb{E}}
\newcommand{\EE}[1]{\E\lrbra{#1}}
\newcommand{\var}{\mathrm{Var}}
\newcommand{\tsum}{\sum\nolimits}
\newcommand{\trace}{\mathrm{tr}}
\newcommand{\ind}[1]{\mathbb{I}\!\lrset{#1}}
\newcommand{\non}{\nonumber}
\newcommand{\poly}{\mathrm{poly}}

\newcommand{\tr}{^{\mkern-1.5mu\mathsf{T}}}
\newcommand{\pinv}{^{\dagger}}
\newcommand{\st}{\star}
\renewcommand{\dag}{\dagger}

\newcommand{\tv}[2]{\mathsf{D}_{\mathrm{TV}}(#1,#2)}
\newcommand{\ent}{\mathsf{H}}
\newcommand{\info}{\mathsf{I}}

\newcommand{\reals}{\mathbb{R}}
\newcommand{\eps}{\varepsilon}
\newcommand{\ep}{\varepsilon}
\newcommand{\sig}{\sigma}
\newcommand{\del}{\delta}
\newcommand{\Del}{\Delta}
\newcommand{\lam}{\lambda}
\newcommand{\half}{\frac{1}{2}}
\newcommand{\thalf}{\tfrac{1}{2}}

\newcommand{\eqdef}{:=}
\newcommand{\eq}{~=~}
\renewcommand{\leq}{~\le~}
\renewcommand{\geq}{~\ge~}
\newcommand{\lt}{~<~}
\newcommand{\gt}{~>~}
\newcommand{\then}{\quad\Rightarrow\quad}

\let\oldtfrac\tfrac

\let\nablaold\nabla
\renewcommand{\nabla}{\nablaold\mkern-2.5mu}

\newcommand{\nf}[2]{\nicefrac{#1}{#2}}
\newcommand{\vct}[1]{\bm{#1}}
\newcommand{\mat}[1]{\bm{#1}}
\newcommand{\ten}[1]{\mathcal{#1}}
\newcommand{\frob}[1]{\|#1\|_2}
\newcommand{\p}[2]{\textrm{Pr}[ #1\in #2] }
\newcommand{\pe}[2]{\mathop{\textrm{Pr}}\limits_{#1}\left[#2\right] }
\newcommand{\inp}[2]{\left\langle #1,#2\right\rangle}
\newcommand{\beps}{\bm{\eps}}
\newcommand{\mX}{\mathcal{X}}
\newcommand{\mY}{\mathcal{Y}}
\newcommand{\mF}{\mathcal{F}}
\newcommand{\RE}[3]{\E{#1}{\sup_{#2}#3}}

\title{Is Your Imitation Learning Policy Better than Mine? 
 \\ Policy Comparison with Near-Optimal Stopping}

 \author{\authorblockN{David Snyder$^{1, 2}$, Asher James Hancock$^{2}$, Apurva Badithela$^{2}$, Emma Dixon$^{1}$, Patrick Miller$^{1}$,\\ Rares Andrei Ambrus$^{1}$, Anirudha Majumdar$^{2}$, Masha Itkina$^{1}$, and Haruki Nishimura$^{1}$}
\authorblockA{$^{1}$Toyota Research Institute (TRI), $^{2}$Princeton University\\ \texttt{dasnyder@princeton.edu}}
}

\maketitle

\begin{figure*}[htbp] 
    \centering
    \includegraphics[width=
    \linewidth]{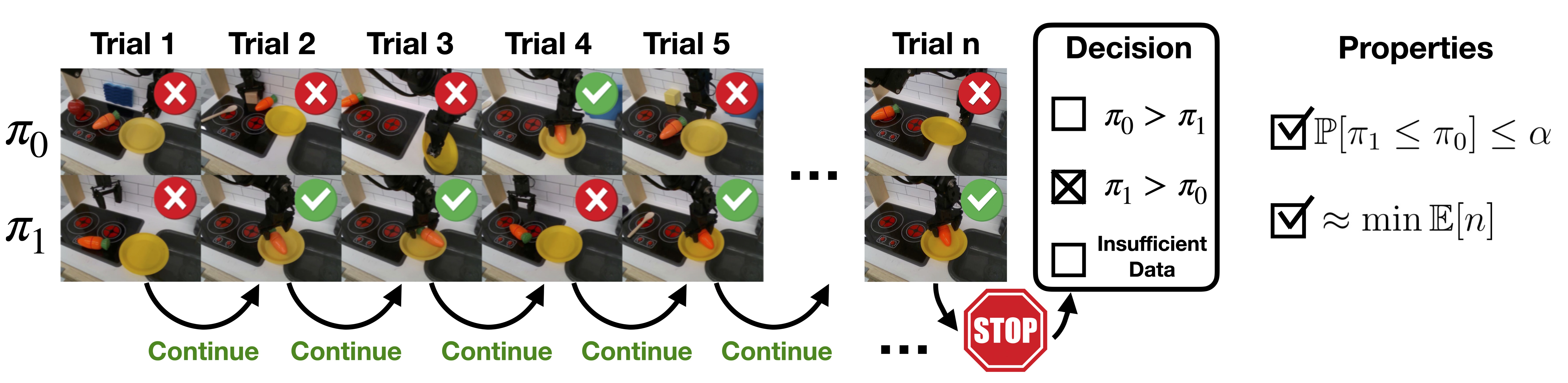}
    \caption{Robot policy comparison problem under binary success/failure metrics. Novel policy $\pi_1$ is compared against baseline $\pi_0$ in a sequence of trials. Within a given evaluation budget, the evaluator seeks a statistically significant comparison in as few trials as possible. Due to the cost of hardware setup and calibration, as well as the limited inference speed of complex policies, these results generally arrive in sequence from a single (or few) hardware setups. Allowing the evaluator to adaptively and near-optimally tailor the number of trials based on the data observed so far --- \emph{without compromising statistical assurances of the comparison} --- is a central contribution of this work.}
    \label{fig:Anchor}
\end{figure*}

\begin{abstract}
Imitation learning has enabled robots to perform complex, long-horizon tasks in challenging dexterous manipulation settings. 
As new methods are developed, they must be rigorously evaluated and compared against corresponding baselines through repeated evaluation trials. However, policy comparison is fundamentally constrained by a small feasible sample size (e.g., 10 or 50) due to significant human effort and limited inference throughput of policies.
This paper proposes a novel statistical framework for rigorously comparing two policies in the small sample size regime.
Prior work in statistical policy comparison relies on batch testing, which requires a fixed, pre-determined number of trials and lacks flexibility in adapting the sample size to the observed evaluation data. Furthermore, extending the test with additional trials risks inducing inadvertent p-hacking, undermining statistical assurances.
In contrast, our proposed statistical test is \textit{sequential}, allowing researchers to decide whether or not to run more trials based on intermediate results. This adaptively tailors the number of trials to the difficulty of the underlying comparison, saving significant time and effort without sacrificing probabilistic correctness. 
Extensive numerical simulation and real-world robot manipulation experiments show that our test achieves near-optimal stopping, letting researchers stop evaluation and make a decision in a near-minimal number of trials. Specifically, it reduces the number of evaluation trials by up to 32\% as compared to state-of-the-art baselines, while preserving the probabilistic correctness and statistical power of the comparison. Moreover, our method is strongest in the most challenging comparison instances (requiring the most evaluation trials); in a multi-task comparison scenario, we save the evaluator more than 160 simulation rollouts.
    
\end{abstract}

\IEEEpeerreviewmaketitle

\section{Introduction}
The importance of trustworthy and efficient robot policy evaluation protocols has become paramount in imitation learning as the scale of underlying deep learning models and the complexity of tasks continue to increase.
This need is especially pronounced in dexterous manipulation where stochastic, contact-rich interactions between the robot and the environment introduce inherent randomness in outcomes.

A particularly important aspect of policy evaluation is \textit{policy comparison}, where two policies are repeatedly deployed in an environment to assess their relative performance. Policy comparison forms the foundation of robot learning as ``an empirical science"~\cite{kress-gazit_robot_2024}, enabling researchers to objectively measure the scientific progress of the field.
Nevertheless, this setting introduces an additional source of stochasticity due to random outcomes of both the first and the second policies, making the reliability of the comparison more challenging to ensure than evaluation of single policies.

To motivate concretely, consider an example policy comparison scenario presented in \cref{fig:Anchor} where the performance is quantified based on binary success/failure metrics, a common choice~\cite{brohan2023rt, xiao2023robot, o2023open} as continuous-valued rewards are often difficult to define. This scenario naturally arises when researchers want to demonstrate the effectiveness of a particular intervention (e.g., a new policy architecture) by comparing the new policy $\pi_1$ against a baseline $\pi_0$. Alternatively, $\pi_1$ and $\pi_0$ could represent the same policy evaluated under different environment distributions, providing insights on generalization. In either case, the evaluator faces two practical challenges. First, only a small number of trials (e.g., 10--60)~\cite{mandlekar2021matters, florence2022implicit, chi2023diffusion, octo_2023, kim2024openvla, black2024pi_0} can be performed per evaluation setting due to the large human effort needed to reset the environment between trials and the substantial wall-clock time imposed by limited inference throughput of large policies. While high-fidelity simulators can alleviate the human effort and still provide valuable insights into policy performance~\cite{li2024evaluating, pumacay2024colosseum}, real-world evaluation remains indispensable for ensuring reliable deployment in downstream applications. Second, the evaluation results are revealed sequentially, possibly leading to fluctuating observations depending on when the evaluation is stopped. In the \cref{fig:Anchor} example, the evaluator could observe more successes for $\pi_0$ after conducting additional trials, even though $\pi_1$ initially appeared superior after the first five.

Although recent work~\cite{Vincent_2024, kress-gazit_robot_2024} proposes statistical policy comparison approaches, it follows the conventional batch testing scheme, requiring complete results from a pre-determined number of trials before the statistical test can be performed. Furthermore, the test can be conducted only once for the corresponding results; even if the test fails to determine the relative performance due to closely matched results, the evaluator cannot append more evaluation trials to the existing results to run the test again, as doing so would constitute p-hacking that invalidates statistical assurances~\cite{ramdas_game-theoretic_2023}. Unfortunately, this is a common but harmful research practice outside of robotics \cite{john2012measuring}, which needs to be averted to ensure reproducible science.

To address these challenges, we propose a novel sequential testing framework named \OurMethod~(\textbf{S}equential \textbf{T}esting for \textbf{E}fficient \textbf{P}olicy Comparison) for rigorously comparing performance of imitation learning policies\footnote{Although this paper focuses on imitation learning, \OurMethod~is naturally applicable to evaluating any types of policies based on binary metrics, including reinforcement learning (RL) policies with sparse 0/1 reward.}. Unlike batch testing, our approach allows the number of trials to be varied within a given experimental budget. This critical feature offers two practical advantages. First, the evaluator can stop conducting trials early without sacrificing the probabilistic correctness of the comparison if enough statistical evidence is accumulated quickly in favor of $\pi_1$ (or $\pi_0$). Second, it reduces the epistemic risk of overconfident (and potentially incorrect) conclusions when $\pi_0$ and $\pi_1$ are closely matched, since the test will abstain from making a decision if statistical evidence remains low after all the planned trials have been executed. Alternatively, the evaluator may safely append additional trials to the original samples and re-conduct statistical analysis on all the results obtained thus far without inadvertent p-hacking. We demonstrate these advantages through simulation and real-world robot manipulation experiments. Furthermore, extensive numerical experiments show that  \OurMethod~significantly outperforms state-of-the-art (SOTA) sequential methods, reducing the required number of trials by up to 32\% without sacrificing probabilistic correctness. The specific contributions of this paper are as follows:
\begin{itemize}
\item We propose \OurMethod, a novel sequential statistical framework for evaluating relative performance of two policies with tunable probabilistic correctness \footnote{Associated website and code can be found at: \url{https://tri-ml.github.io/step/}}.
\item Our sequential testing approach admits an adaptive number of evaluation trials tailored to the difficulty of the underlying comparison while achieving near-minimal sample complexity.
\item We additionally present a straightforward extension of our framework to (1) multi-task and (2) multi-policy comparison settings via a reduction to a set of pair-wise statistical comparisons.
\end{itemize}

\section{Preliminaries}
\label{sec:preliminaries}
Consider a physical robot trained to complete a task in a variety of environment realizations. This setting is naturally modeled as a partially observable Markov decision process (POMDP)~\cite{kaelbling1998planning}, where the underlying state $s$ represents the environment and robot states. The observation $o$ is determined by the robot's embodiment and sensing apparatus. The training pipeline is designed to synthesize a policy taking actions $a$ that achieve a high reward $r(s, a) = \mathbbm{1}[s \in \mathcal{S}_\text{success}]$ on a particular task, where $\mathcal{S}_\text{success}$ is a set of successful terminal states in which the episode will terminate. This reward encodes a binary success/failure criterion of the task. In imitation learning, a surrogate loss function is used to train a policy that matches the behavior of expert demonstrations \cite{ross2011reduction} instead of directly solving the POMDP.
At the end of a training process, a policy $\pi_1$ will be obtained. This policy has a true success rate (i.e., expected total episode reward) under a distribution $\mathcal{D}_{s_0, o_0}$ over the initial state and observation:
\begin{equation}
    p_1 = \mathbb{E}_{\mathcal{D}_{s_0, o_0}, \pi_1} \bigg[\sum_{t = 0}^T r(s_t, a_t)\bigg],
\end{equation}
where dependence on the state transition and observation models are omitted for brevity. The true success rate is \emph{unknown} and must be estimated via multiple evaluation trials.
\begin{assumption}
    [Regularity] In each evaluation trial, the initial state $s_0$ and the observation $o_0$ are drawn in an independent and identically distributed (i.i.d.) fashion from the underlying distribution $\mathcal{D}_{s_0, o_0}$\footnote{Note that this is a standard assumption in statistical testing. A discussion of practical methods by which to approximately satisfy this condition in robotic evaluation is included in \cite{kress-gazit_robot_2024}.}. We assume access to samples from $\mathcal{D}_{s_0, o_0}$, but do not assume $\mathcal{D}_{s_0, o_0}$ itself to be known.
\end{assumption}

Under Assumption 1, the $n^{\text{th}}$ evaluation trial involves making an i.i.d. draw of an environment from $\mathcal{D}_{s_0, o_0}$ and running the policy $\pi_1$ in this environment. This yields a binary success/failure outcome $z_{1, n}$ corresponding to an i.i.d. draw from a Bernoulli random variable with mean $p_1$, which is the true performance (success rate) of the policy $\pi_1$ on the task: 
\begin{equation}
    z_{1, n} \sim \text{Ber}(p_1).
\end{equation}
Here, $z_{1, n} = 1$ indicates success and $z_{1, n} = 0$  failure. 
For any baseline policy $\pi_0$, we similarly denote the outcome of its $n^\text{th}$ trial as $z_{0, n} \sim \text{Ber}(p_0)$. 
For the sake of statistical analysis, we pair the outcomes of two policies by their indices in a vector $Z_n = (z_{0, n}, z_{1, n})$.
The policy comparison problem can be formalized in the sense of Neyman-Pearson (N-P) statistical testing \cite{neyman_1933}, where the null (skeptical) hypothesis is that the novel policy $\pi_1$ is no better than the baseline $\pi_0$ and the alternative is that the novel policy is indeed better:
\begin{equation}
    \label{Eqn:Test}
    \begin{split}
        \text{Null Hypothesis } \mathbb{H}_0: & \hspace{2mm} p_1 \leq p_0 \equiv (p_0, p_1) \in \mathcal{H}_0\\
        \text{Alt. Hypothesis } \mathbb{H}_1: & \hspace{2mm} p_1 \gt p_0 \equiv (p_0, p_1) \in \mathcal{H}_1. 
    \end{split}
\end{equation} 

\begin{figure}[!thb]
    \centering
    \includegraphics[width=0.45
    \textwidth]{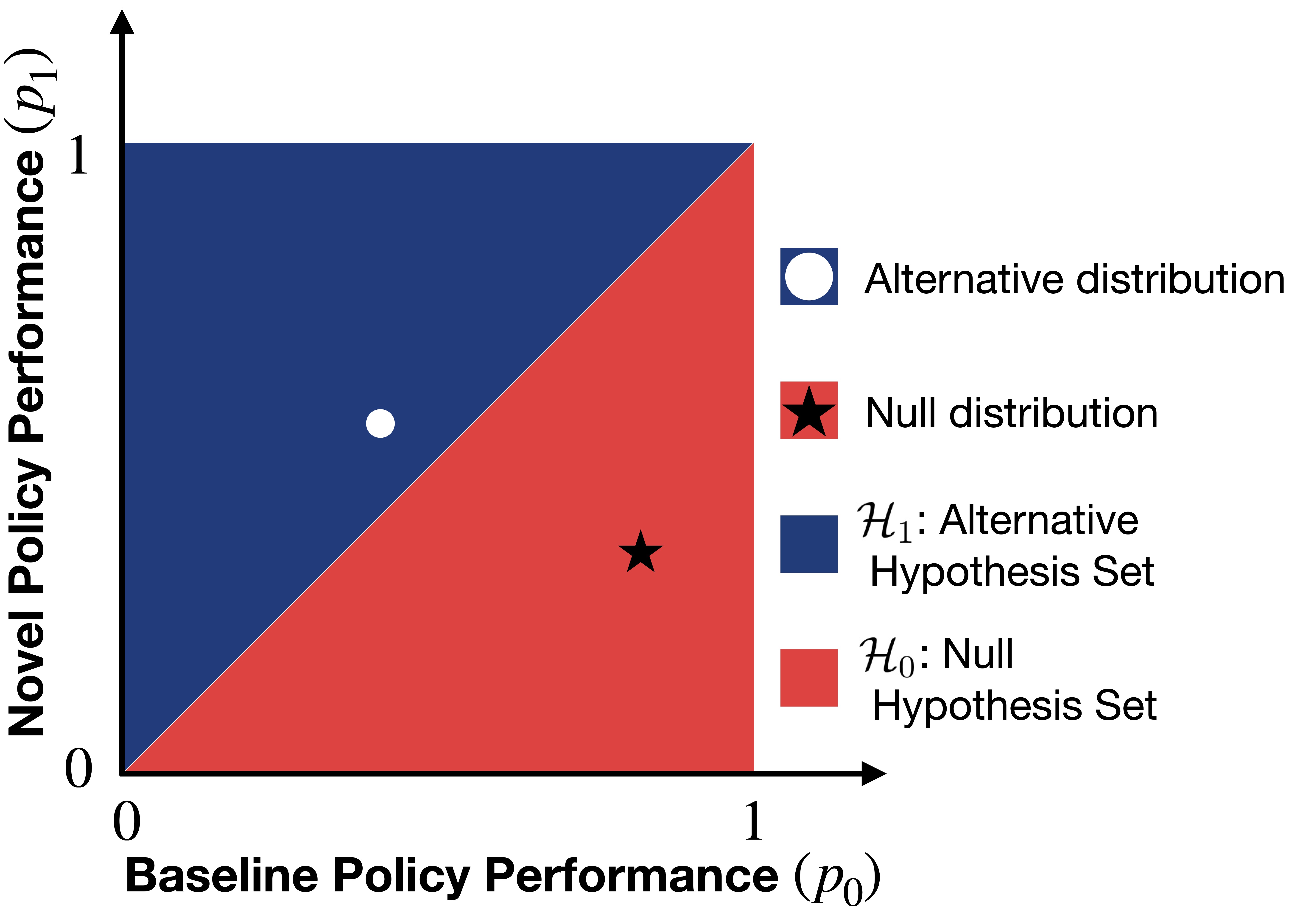}
    \caption{The policy comparison problem as a composite-vs-composite statistical test. The null hypothesis set (red) corresponds to the novel policy being worse ($p_1\leq p_0$), while the alternative hypothesis set (blue) corresponds to the novel policy being better ($p_1 > p_0$). The sets are each termed ``composite" because they contain many elements. For any pair of policies, the truth corresponds to a single point; as examples, the white circle is a case where the alternative is true (baseline success 45\%, novel policy success 55\%), while the black star corresponds to the null being true (baseline success 82\%, novel policy success 35\%). }
    \label{fig:ComparisonTestSetting}
\end{figure}

As illustrated in \cref{fig:ComparisonTestSetting}, this amounts to a composite vs. composite statistical test \cite{bickel_mathematical_2015}. A hypothesis is termed ``simple" if it is singleton, i.e., corresponds to a single data-generating distribution. For example, $(p_0, p_1) = (0.3, 0.7)$ would be a simple hypothesis. If the hypothesis corresponds to multiple data-generating distributions, it is termed ``composite".

The Type-I error rate (denoted $\alpha$) of a statistical test corresponds to the probability of falsely rejecting the null \emph{under the worst-case singleton element $h_0$ of the null hypothesis set $\mathcal{H}_0$} -- that is, under the hardest-to-distinguish $(p_0, p_1) \in \mathcal{H}_0$. This type of error corresponds to falsely concluding that $\pi_1$ is better than $\pi_0$ when it is not. The \emph{power} of a test (denoted $1-\beta$) is the mixture probability of correctly rejecting the null when the alternative is true, under some measure on the set of alternatives\footnote{For example, a minimax measure mimics the Type-I setting (e.g., the worst-case singleton $h_1 \in \mathcal{H}_1$). However, due to the finite termination condition, such a measure is vacuous because there will always exist a $(p_0, p_0+\epsilon)$ for $\epsilon > 0$ sufficiently small such that the attainable power will only negligibly exceed random guessing. Therefore, in practice other measures must be used.}. The Type-II error $\beta$ is the associated (mixture) probability of failing to reject the null when the alternative is true. This represents failing to conclude that $\pi_1$ is better, when it is in fact better than $\pi_0$. 

Finally, while the preceding development has focused on one-sided testing, the comparison problem naturally admits a bidirectional version allowing decisions for membership in either the alternative set (\textbf{RejectNull}) or the null set (\textbf{AcceptNull}). Note that the decision \textbf{AcceptNull} formally amounts to rejecting the null of a ``flipped" version of \cref{Eqn:Test}. In all subsequent discussion we will implicitly utilize the bidirectional test, which will allow for the decision \textbf{AcceptNull}. 

\section{Problem Formulation}
\label{sec:problem}

We assume that a robot evaluator is tasked with distinguishing two policies via successive evaluations, resulting in the testing paradigm described in \cref{sec:preliminaries}. We also assume that the evaluator has \emph{pre-selected} the desired significance level $\alpha^*$ of the comparison and a maximum number of trials (for each policy) that they are willing or able to run: $N_\text{max}$. 

First, we note that the Type-I error must always be controlled in statistical testing, i.e., upper-bounded at the evaluator-specified rate $\alpha^* \in (0, 1)$. This represents a hard constraint (``validity") that \emph{will be enforced in all subsequent testing procedures}; a test is not feasible if it is not Type-I error controlling. The evaluator's goal is to synthesize a decision rule that limits the Type-I Error to $\alpha^*$ while maximizing power and minimizing the expected number of evaluation trials. 

We assume an underlying representation of the evaluation information collected \emph{thus far} is available to the evaluator in a state $x_n = F(Z_1, Z_2, ..., Z_n)$. Then, the evaluator's procedure consists of deciding whether to \textbf{Continue} (gather another trial for each policy) or stop (and either \textbf{AcceptNull} or \textbf{RejectNull}). Given a decision set $\mathcal{U} = \{\textbf{AcceptNull}, \textbf{Continue}, \textbf{RejectNull}\}$, the problem is to find a state partition $\zeta = u(x)$ to optimally balance minimizing the expected sample size and maintaining high power, conditioned on the Type-I Error rate constraint, as shown in \cref{Eqn:PracticalOpt}:
\begin{equation}
    \label{Eqn:PracticalOpt}
    \begin{split}
    \min_{\zeta: \mathcal{X} \mapsto \mathcal{U}} & \hspace{2mm} \mathbb{E}_{\mu(\mathcal{H}_1)}[n_{\text{stop}} + c\beta_{N_{\max}}] \\
    \text{s.t. } & \hspace{2mm} \max_{h_0 = (p_0,\text{ }p_1) \in \mathcal{H}_0} \alpha(\zeta, h_0) \leq \alpha^* \\
    & \hspace{2mm} 0 \leq n_{\text{stop}} \leq N_{\text{max}} \text{ w.p. 1}.
    \end{split}
\end{equation}
This function is a multi-objective optimization which seeks to simultaneously minimize expected sample size and maximize power subject to the validity constraint. Informally, for any feasible terminal power $1-\beta_{N_{\max}}^{\text{feasible}}$ (conditioned on $N_{\text{max}}, \alpha^*$, and the true underlying distribution, which we emphasize is \emph{unknown a priori}), there is a value of $c > 0$ that effects a decision rule in \cref{Eqn:PracticalOpt} approximating a test controlling $\beta =\beta_{\text{feasible}}$. For example, this framework recovers the batch problem as $c \rightarrow \infty$ (demanding maximal power), and immediate termination as $c \rightarrow 0^+$\footnote{For the latter case: without looking at any data, draw a random number uniformly on [0, 1]. If it is less than $\alpha^*$ reject the null, otherwise fail to reject and terminate. This terminates at step 0 with probability 1, has power $\alpha^*$, and is valid.\label{footnote}} (demanding minimal sample complexity).
This objective will govern the methodology and analysis presented in the rest of the paper. 

\section{Methodology}
\label{sec:methodology}
There are several important practical considerations to constructing a near-optimal solution to \cref{Eqn:PracticalOpt}. We present the concrete challenges first, and then discuss the technical innovations that account for them. Throughout, we will leverage significant mathematical structure in the testing problem. Where insightful or intuitive, this will be explained \emph{in situ}. See \cref{subsec:Math} in the Supplement for additional details.

\subsection{State Representation}
First, we express the information available to the evaluator at time $n$ (i.e., after $n$ evaluation trials have been performed for each policy) in a control-theoretic state representation; the evaluator's decision can then be understood as a state-feedback decision rule. In selecting this representation, the first instance of mathematical structure is the membership of Bernoulli distributions in the univariate exponential family; such distributions have known sufficient statistics which represent (informally) an optimal compression of the data for the purposes of testing and estimation \cite{bickel_mathematical_2015}. Thus, a natural (``near-minimal") state representation is precisely the element-wise sufficient statistic for $p_0$ and $p_1$ respectively, augmented with a time state. For univariate exponential family distributions, the sufficient statistic is the sum of the observed data (i.e., respective number of successful trials under $\pi_0$ and $\pi_1$); this makes the state representation a first-order discrete integrator, as shown in \cref{Eqn:dynamics}: 
\begin{equation}
    \label{Eqn:dynamics}
    \begin{split}
    \mathbf{x}_{n+1} & = \mathbf{x}_n + \mathbf{d}_n \\
    \mathbf{d}_n &= (z_{0,n}, z_{1,n}, 1) \sim (\text{Ber}(p_0),\text{ }\text{Ber}(p_1),\text{ }1) \\
    \mathbf{x}_0 & = \mathbf{0}_3.
    \end{split}
\end{equation}
Unlike in a typical control problem, we cannot actively guide the state trajectory; we are instead deciding when to stop based on the state trajectory. Concretely, the ``control" involves partitioning the state space into stopping and continuation regions. In robotics, a similar set-theoretic notion arises in robust control and safe navigation, through the generation of invariant sets for dynamic systems (i.e., in reachability and barrier function-type methods \cite{stipanovic_computation_2003, ames_control_2019}), though those methods are primarily interested in nonstochastic uncertainty. Separately, many applications of asset pricing in mathematical finance utilize optimal stopping to set options prices under stochastic uncertainty \cite{myneni_pricing_1992}. 

\subsection{Decision Regions}
\label{subsec:DecisionRegions}
In the sequential problem the state space partitions occur in sequence, yielding an effective representation of the decision rule $\zeta: \mathcal{X} \mapsto \mathcal{U}$ in the form of \cref{Eqn:decisionregion}:
\begin{equation}
\label{Eqn:decisionregion}
    \zeta \equiv \bigg\{\mathcal{X}^{\text{Reject Null}}_{n}, \mathcal{X}^{\text{Accept Null}}_{n}, \mathcal{X}^{\text{Continue}}_{n}\bigg\}_{n=1}^{N_\text{max}}.
\end{equation}
At each time step, the sets jointly encode control decisions for every state.
Intuitively, the larger the size of the rejection region $\mathcal{X}_{n}^{\text{Reject Null}}$ (resp. $\mathcal{X}_{n}^{\text{Accept Null}}$), the smaller the number of expected trials needed to reject the null (resp. alternative), achieving a lower value of the stopping time component of \cref{Eqn:PracticalOpt}. Focusing on $\mathcal{X}_{n}^{\text{Reject Null}}$ in the following development\footnote{As discussed in \cref{sec:preliminaries}, the case of rejecting the alternative and accepting the null can be considered by flipping $p_0$ and $p_1$.}, our auxiliary objective is then to maximize the size of the set $\mathcal{X}_{n}^{\text{Reject Null}}$ globally across all $n \in \{1,\cdots,N_{\max}\}$.
We take a probabilistic approach, allowing the states to reject with a probability less than 1. As we will see in \cref{subsec:Optimization}, this yields an efficient optimization problem.

In addition to the maximization, there are two core challenges to synthesizing such a partition. First, the Type-I Error rate must be controlled. The decision to stop and reject the null hypothesis in a state $\mathbf{x}_n$ incurs risk due to the probability that $\mathbf{x}_n$ was reached under data generated from some $(p_0, p_1) \in \mathcal{H}_0$ -- i.e., under the null hypothesis in \cref{Eqn:Test}. Thus, having a large $\mathcal{X}^{\text{Reject Null}}_{n}$ risks violating Type-I Error rate control. Second, the temporal rate at which the risk is accrued must be set appropriately to encourage early stopping on easy instances (i.e., $\pi_1$ significantly outperforms $\pi_0$) without harming performance too much on harder instances (i.e., $\pi_1$ and $\pi_0$ are closely competing, which requires many trials to distinguish). We will first address Type-I Error control in \cref{subsec:Method_Type1}, and the temporal risk accumulation in \cref{subsec:RiskBudgets}. The resulting tractable optimization problem is presented in \cref{subsec:Optimization}.

\subsection{Type-I Error Control}
\label{subsec:Method_Type1}
The composite nature of the null hypothesis (which contains all pairs $\mathcal{H}_0 = \{(p_0, p_1) \in [0, 1]^2 \text{ }\rvert \text{ }p_0 \geq p_1\}$) means that the decision-making problem can be thought of (informally) as analogous to distributionally robust control, where the uncertainty is over the particular worst-case $(p_0, p_1) \in \mathcal{H}_0$. Controlling the Type-I Error in policy comparison then amounts to controlling the Type-I Error uniformly (robustly) for all $h \in \mathcal{H}_0$. Suppose that some rejection region for the first $n - 1$ steps has been obtained with accumulated risk $\alpha_{n-1}$, and we are interested in bounding the Type-I Error for $n$ by some $\alpha_n > \alpha_{n-1}$. Given the notion of stopping and continuation regions introduced in the previous section, this is expressed mathematically as: 
\begin{equation}
    \label{Eqn:Type1Control}
    \max_{h \in \mathcal{H}_0} \mathbb{P}_{h}\left(\mathbf{x}_n \in \mathcal{X}^{\text{Reject Null}}_{n} \mid  \mathcal{X}_{n - 1}^{\text{Reject Null}} \right) \leq \alpha_n.
\end{equation}

The dependence on the (probabilistic) rejection region for $n - 1$ is made explicit in \cref{Eqn:Type1Control}, reflecting the internal dynamic structure. For example, if some state $\mathbf{x}_{n-1} = (a, b, n - 1)$ is rejected at $n - 1$ with a non-zero probability, then the 1-step reachable states (e.g., $(a + 1, b, n)$) under the dynamics \cref{Eqn:dynamics} are less likely to be feasible at $n$. 
The presence of many (infinite) elements $h$ in the set of null hypotheses $\mathcal{H}_0$ makes verification of Type-I Error control challenging \textit{a priori}, even for a single particular $n$. However: monotonicity, symmetry, and smoothness properties of this family of testing problems (noted in, for example, \cite{Chernoff_1986}) allow for efficient discretization procedures that preserve safety.
Specifically, it suffices to consider a discrete set of the ``worst-case" nulls $\hat{\mathcal{H}}_0 = \{(p^{(1)}, p^{(1)}), (p^{(2)}, p^{(2)}), \cdots, (p^{(M)}, p^{(M)})\}$, where $0 \leq p^{(1)} \cdots \leq p^{(M)} \leq 1$. Further details of this discretization process is given in \cref{subsubsec:null_discretization} in the Supplement. Thus, \cref{Eqn:Type1Control} reduces to a finite set of individual inequality constraints.

We can equivalently represent this set of constraints as a linear inequality $P_n\mathbf{w}_{n} \leq \alpha_n \mathbbm{1}$, where $\mathbf{w}_{n}$ is a vector representing the probability of rejecting the null in each state $\mathbf{x}_n \in \mathcal{X}_{n} = \{(0, 0, n), (0, 1, n), \cdots (N_{\max}, N_{\max}, n)\}$. $P_n$ is a non-negative matrix of size $(M, |\mathcal{X}_{n}|)$ where each row represents the probability of reaching particular states under $h^{(i)} = (p^{(i)}, p^{(i)})$:
\begin{equation}
    \label{Eqn:state_occupancy}
    (P_n)_{ij} = \mathbb{P}_{h^{(i)}}\left(\mathbf{x}_n = \mathbf{x}_n^{j} \mid \mathcal{X}_{n - 1}^{\text{Reject Null}} \right),
\end{equation}
where $\mathbf{x}_n^j$ is the $j^\text{th}$ (discrete) state in $\mathcal{X}_{n}$.
Given the rejection region from the previous time step $n - 1$, we can accurately compute this probability \cref{Eqn:state_occupancy} by forward-propagating the previous state occupancy distribution $(P_{n - 1})_{i}$ according to the stochastic dynamics model \cref{Eqn:dynamics} under $h^{(i)}$.

\subsection{Power Adaptivity to Varying Difficulty: Risk Budgets}
\label{subsec:RiskBudgets}
As discussed in \cref{subsec:DecisionRegions}, we must appropriately adjust the temporal rate of risk accumulation.
To formally define this notion of risk, we introduce a non-negative scalar function $f(n)$ for $n \in \{1, \cdots, N_{\max}\}$, which determines the maximum allowable Type-I Error under any null hypothesis at each step $n$. We impose a constraint $\sum_{n=0}^{N_\text{max}} f(n) = \alpha^*$ to globally bound the Type-I error by $\alpha^*$.
This risk ``budget" can be interpreted as encoding the evaluator's competing objectives: to reject the null and stop the evaluation quickly in easier cases (front-loading the risk accumulation) against the desire to wait longer to achieve a significant decision in harder instances (delaying risk accumulation until more data is collected). Importantly, \emph{any risk budget $f$ that is nonnegative everywhere and sums to some $r \in [0, \alpha^*]$ maintains Type-I error control at level $\alpha^*$}. However, the shape of the risk budget will significantly influence the power of the resulting procedure, and thus represents an important component in solving (near-optimally) \cref{Eqn:PracticalOpt}. With this said, in the following experiments we fix the risk budget to be uniform in order to focus on optimizing the decision regions (described in \cref{subsec:Optimization}); to be explicit: this means the risk budget is $f(n)=\frac{\alpha^*}{N_{\text{max}}}$ for each scenario. This selection is heuristically reasonable, but leaves potential for further improvements in future work.

\subsection{Tractable Optimization}
\label{subsec:Optimization}
Having specified the risk budget $f(n)$, it is straightforward to verify that the Type-I Error control is achieved if the following constraint is satisfied for all $n \in \{1, \cdots, N_{\max}\}$:
\begin{equation}
     P_n\mathbf{w}_{n} \leq \sum_{k = 1}^{n} f(k) \mathbbm{1}
\end{equation}
Under this constraint, our objective is to maximize the size of the rejection region. We propose to solve the following series of optimization problems to tractably construct the rejection regions, one for each $n$: 
\begin{equation}
    \label{Eqn:LinearProgram}
    \begin{split}
        \max_{\lVert \mathbf{w}_{n} \rVert_\infty \leq 1} & \hspace{3mm} \mathbbm{1}^T\mathbf{w}_{n} \\
        \text{s.t. } & \hspace{3mm} P_t\mathbf{w}_{n} \leq \sum_{k=1}^n f(k) \mathbbm{1} \\
        \text{ } & \hspace{3mm} 0 \leq \mathbf{w}_{n} \leq 1,
    \end{split}
\end{equation}
This objective encourages the rejection from as many states as possible, maximizing the size of $\mathcal{X}_{n}^{\text{Reject Null}}$. Furthermore, it implicitly rejects from states less likely to occur under any null hypothesis, which are ``cheaper" in terms of accruing risk.
The first constraint ensures that the Type-I error is controlled up to time $n$, as discussed in \cref{subsec:Method_Type1} and \cref{subsec:RiskBudgets}. The second constraint is to enforce boundedness of rejection probabilities in $[0, 1]$.

\begin{algorithm}[t]
\caption{STEP Decision-Rule Synthesis}\label{alg:step}
\begin{algorithmic}
\STATE \textbf{Input: } $N_\text{max} > 0$, risk budget function $f(n)$, type-I error limit $\alpha^* \in (0, 1)$, number of approximation points $M$
\STATE \textbf{Initialize: } $\zeta_0 = \emptyset$, $(P_0)_{ij} = 1$ if $(i, j) = (0, 0)$ else $0$.
\FOR {n $\in \{1, ..., N_\text{max}\}$}
\STATE $P_n \gets \textbf{Propagate}(P_{n - 1}, \zeta_{n-1}, M)$  \hfill \COMMENT{\cref{Eqn:state_occupancy}}
\STATE $\mathbf{w}_n \gets \textbf{Opt}(P_n, f)$ \hfill \COMMENT{\cref{Eqn:LinearProgram}}
\STATE $\zeta_n \gets \textbf{Compress}(\mathbf{w}_n)$
\ENDFOR
\RETURN $\zeta = \{\zeta_1, \dots, \zeta_{N_\text{max}}\}$ \hfill \COMMENT{STEP policy} \hspace{1.5mm}
\end{algorithmic}
\end{algorithm}

The optimization problem \cref{Eqn:LinearProgram} is a simple linear program that can be efficiently solved by a standard optimization software; it is a maximization of $\lVert \mathbf{w}_{n} \rVert_1$ over the nonnegative orthant, subject to additional linear inequality constraints to control Type-I Error. Because each $P_n$ depends on the corresponding $\zeta_{n-1}$, the optimization needs to be performed sequentially for each $n$ in increasing order. Nevertheless, all the computation can be performed offline prior to running the actual policy evaluation. See \cref{alg:step} for the decision-rule synthesis procedure. The fact that the optimization problems are sequentially solvable is owed to the isolation of the risk accumulation rate $f(n)$ as a tunable parameter; otherwise, the rejection regions would be generally coupled across $n$ and the optimization would be more challenging. Finally, we note in \cref{alg:step} that the representation of the optimal policy can be compressed substantially from the vector $\mathbf{w}_n$ to a set of connected sets $\zeta_n = \bigg\{\mathcal{X}^{\text{Reject Null}}_{n}, \mathcal{X}^{\text{Accept Null}}_{n}, \mathcal{X}^{\text{Continue}}_{n}\bigg\}$ defined by their boundaries, due primarily to the general monotonicity of the optimal regions and the unimodality of the distribution of the test statistic.

\section{Experiments}
\label{sec:experiments}

\begin{figure*}[htbp!]
    \centering
    \includegraphics[width=\textwidth] {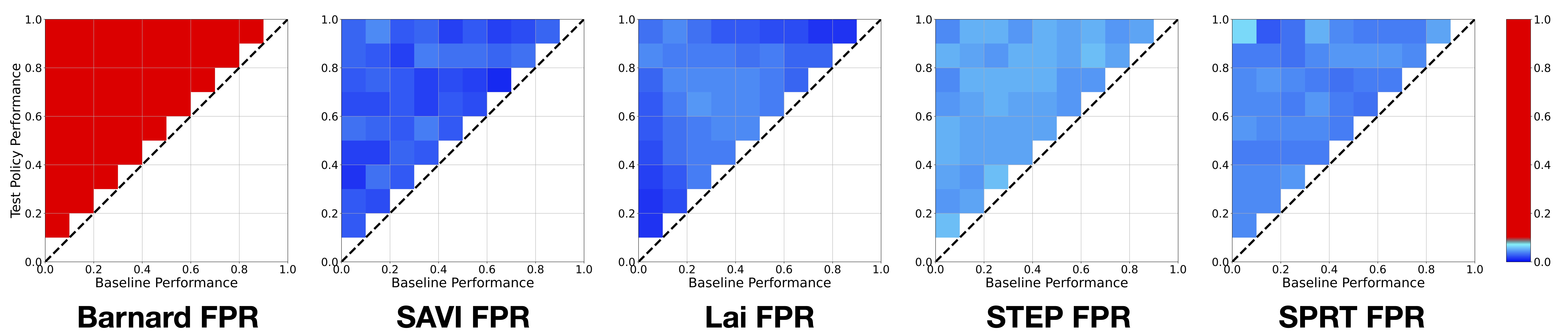} 
    \caption{False positive rate of four feasible methods (Barnard, SAVI, Lai, and Ours (\OurMethod)) and the SPRT (Oracle method) for 400 simulated trajectories drawn from the respective worst-case null distributions for each of 45 alternatives (squares in color); $N_\text{max}=500$ and $\alpha^* = 0.05$. Note that naively utilizing a batch method in sequence leads to violation of Type-1 Error control (red). Additionally, note that SAVI and Lai struggle to utilize the full risk budget in finite $N_\text{max}$ (darker blue regions). }
    \label{fig:terminal_fpr_N500}
\end{figure*}

The experiments are designed to investigate the following key aspects of effective evaluation: 
\begin{enumerate}
\item[A.] How does sequential testing compare with batch testing in terms of statistical validity (Type-1 Error)?
\item[B.] What are the unique advantages of \OurMethod~in sequential comparison problems of varying difficulty? 
\item[C.] How sample efficient is \OurMethod~in practical policy comparison settings? 
\end{enumerate}
The experiments address these questions through extensive numerical validation on simulated success/failure data, as well as practical validation on both simulated rollouts and tasks on physical hardware. 
\subsection{Baseline Procedures}
The baselines constitute the SOTA sequential analysis methods described in \citet{lai_nearly_1988} and \citet{lai_nearly_1994} (termed ``Lai") and the Safe, Anytime-Valid Inference (SAVI) method of \citet{turner_exact_2023}, which is specifically tailored to contingency table (i.e., policy comparison) problems (termed ``SAVI"). 

Similar to \OurMethod, Lai is a valid sequential method under a pre-determined $N_{\max}$ and is asymptotically optimal as $N_{\max}$ tends to infinity. The specific difference between these methods is that Lai makes the asymptotic assumption of normality of the empirical means. This allows for any distribution subject to the central limit theorem (CLT) to be analyzed tractably in the \emph{large-data regime}, as the resulting optimal stopping PDE amounts to the solution of a reverse heat equation. \OurMethod, by contrast, solves the \emph{exact} PDE at the fundamental finite-sample resolution of the testing problem. As will be shown, the Lai procedure does best in difficult (i.e., requiring a lot of data) and symmetric (i.e., $p_0, p_1$ each near $0.5$) test settings, as these best match the asymptotic approximation where the CLT approximation is well-matched to the exact distribution of the empirical means. SAVI, by contrast, does not require a fixed $N_{\max}$ \textit{a priori} and maintains validity for an arbitrarily large $N_{\max}$. In essence, SAVI procedures construct a test statistic subject to stochastic dynamics (the sequential incorporation of data) which possesses a high-probability `stability certificate' under any null data generating process. The `stability' in this case grants time-uniformity (in perpetuity, i.e., not requiring any finite $N_\text{max}$); thus, under any null hypothesis, the probability of \emph{ever} violating the certificate is bounded (by design) to be less than $\alpha$. In exchange for increased generality and robustness, however, SAVI methods inevitably sacrifice sample complexity in cases where a finite upper-bound to $N_{\max}$ is determined by practical time and resource constraints that researchers are subject to. We emphasize that each existing method is widely applicable to practical testing problems beyond robotics, as is STEP. \emph{Importantly, STEP is tailored specifically to the small-$N_\text{max}$ evaluation regime.}

Additionally, an Oracle Sequential Probability Ratio Test (SPRT) \cite{wald_sequential_1945} is run using the true singleton alternative and associated worst-case singleton null; this method represents a near-optimal procedure for easier problems where feasible methods quickly approach a terminal power close to 1 and still serves as a reasonable benchmark in harder problems. \emph{We emphasize that in practical cases, the Oracle method is infeasible to the evaluator;} it is included to give a conservative estimate of the optimality gap of each method. Additional information about each baseline is included in  \cref{sec:related_work}.

\begin{figure*}[!thb]
    \centering
    \includegraphics[width=\textwidth] {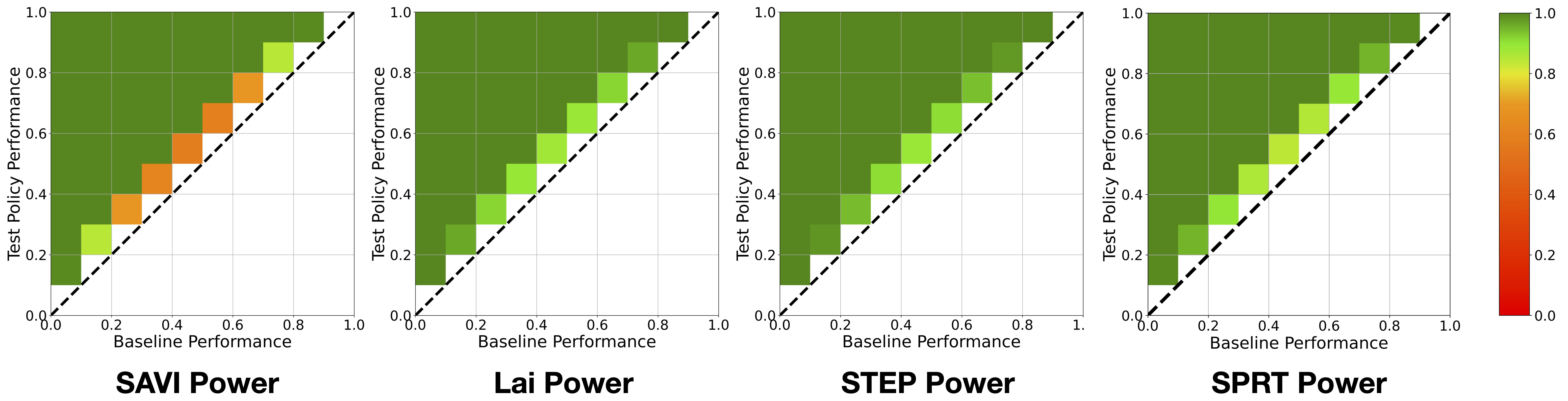} 
    \caption{Terminal power of three feasible methods (SAVI, Lai, and Ours (\OurMethod)) and the SPRT (Oracle method) for 5000 simulated trajectories on each of 45 alternatives (squares in color); $N_\text{max}=500$ and $\alpha^* = 0.05$. Because $N_\text{max}$ is large, the terminal power is high for all but the most difficult cases; however, note that SAVI has significantly worse power in these difficult instances where $p_0$ and $p_1$ are closely competing. This is due to its inherent validity for arbitrary $N_\text{max}$, which is unnecessary in modern robotics evaluation contexts and renders the methods conservative in harder instances.}
    \label{fig:terminalpower}
\end{figure*}

\subsection{Numerical Simulation}

For evaluation, we discretize a grid over the space of alternatives in 10-percentage point increments, offset from zero by five points. There are forty-five resulting alternatives in which $p_1 > p_0$: $(p_0, p_1) = \{(0.05, 0.15), (0.05, 0.25), ..., (0.85, 0.95)\}$. For each of these alternatives, 5000 trajectories (each of length $N_\text{max}$) are simulated from $(\text{Ber}(p_0), \text{Ber}(p_1))$. In addition, 400 additional trajectories are generated under the worst-case null for each of the 45 alternatives, in order to verify the Type-I Error control. This evaluation data is shared across each methodology.

The algorithms are first validated on multiple pairs of $(N_\text{max}, \alpha^*)$. The ensuing numerical results will use the case $N_\text{max} = 500, \alpha^* = 0.05$; similar figures for $N_\text{max}=100, \alpha^* = 0.05$ are included in \cref{subsec:results_Nmax_100} in the Supplement. For each method, we compute for each of the 45 alternatives the following characteristics: (1) Type-I Error -- the fraction of associated worst-case null trajectories which have incorrectly rejected the null; (2) Terminal Power -- the fraction of alternative trajectories which have correctly rejected the null by step $N_\text{max}$; (3) Cumulative Power -- a visualization of the fraction of alternative trajectories which have correctly rejected the null by step $t$ for all $t \in \{0, 1, ..., N_\text{max}\}$. Note that another natural evaluation metric, the expected time-to-decision ($\mathbb{E}[n_\text{stop}]$), can be derived from the cumulative power as the area between the curve and the constant $y = 1$. 
\subsubsection{How does sequential testing compare with batch testing in Type-I Error control?}
We show the Type-1 Error control of each sequential method in \cref{fig:terminal_fpr_N500}, as well as a widely-used batch method (Barnard's Test \cite{barnard_significance_1947}) run in sequence. Each sequential method maintains Type-1 Error control; however, STEP is the most efficient at using the full risk budget (lightest blue), while Lai is weaker (more conservative) in the lower-variance regime and SAVI is weaker in general (darker blue). Using a batch method like Barnard's Test in sequence, conversely, violates Type-1 Error control (red). This highlights the aforementioned p-hacking issue of the batch testing scheme. A rigorous batch evaluator, having chosen $N$, cannot adapt to the data as it arrives lest they invalidate a resulting conclusion. On the other hand, \OurMethod~provides a safety margin for continued testing with the maximum allowable sample size $N_{\max} > N$.
\subsubsection{What are the unique advantages of \OurMethod~in comparison problems of varying difficulty?}
\paragraph{Terminal Power}
We begin a more fine-grained comparison of the efficiency of sequential procedures by presenting the terminal power (probability of deciding \textbf{RejectNull} by step $N_\text{max}$) of all feasible methods and the Oracle SPRT in \cref{fig:terminalpower}. This metric serves to illustrate a downside of SAVI: its inherent validity at arbitrarily large $N_\text{max}$ imposes strong finite-time costs. In this case, the terminal power when the true gap in policy performance is 10 percentage points is significantly lower than the Lai baseline and our \OurMethod, which closely approximate the SPRT Oracle. 

\paragraph{Cumulative Power}
We now demonstrate the downside of the Lai baseline procedure as compared to our method through the cumulative power of the procedure. 

\begin{figure*}[!thb]
    \centering
    \includegraphics[width=\textwidth] {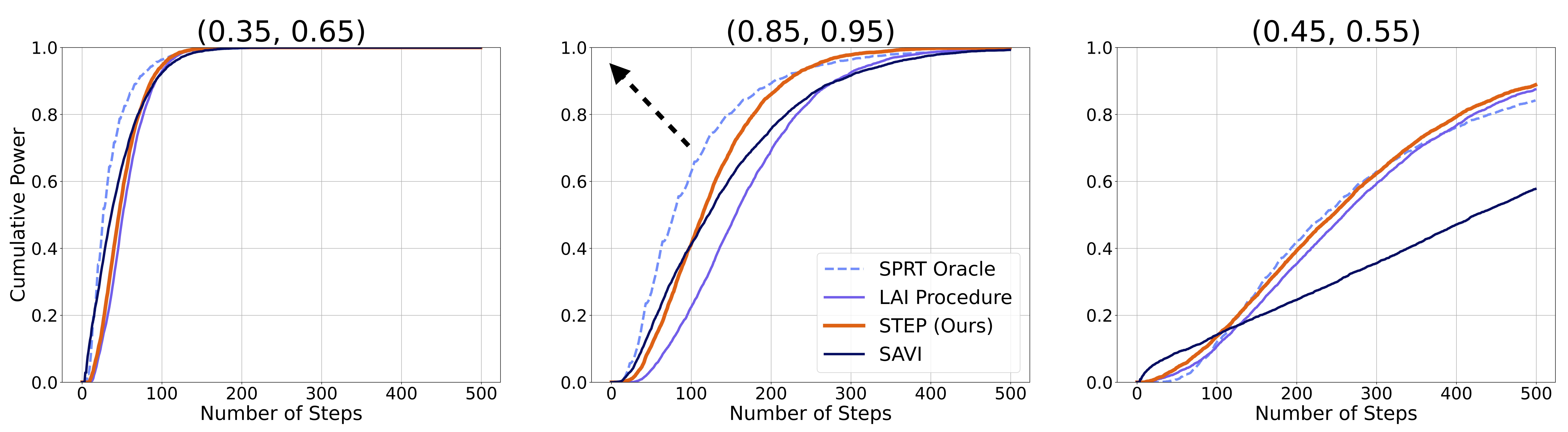}
    \caption{Cumulative power of all feasible methods (Lai, SAVI, STEP (Ours)) and SPRT Oracle over 5000 trajectories in three evaluation settings of increasing difficulty; $(p_0, p_1)$ for each setting title the respective figures. $N_\text{max}=500$ and $\alpha^* = 0.05$. The expected time-to-decision is the integral of the area \emph{above} the cumulative power curve; therefore, curves higher and to the left are better (black arrow). (Left) For a gap of 30 percentage points, all methods demonstrate similar stopping times. (Center) For a gap of 10 percentage points in the low-variance regime (i.e., farther from 0.5), STEP significantly outperforms the Lai procedure and is better than SAVI. (Right) For a gap of 10 percentage points in the high-variance regime, SAVI struggles to maintain high power, and underperforms the other methods. }
    \label{fig:cumulativepower}
\end{figure*}

In \cref{fig:cumulativepower} (right), we illustrate the cumulative power on a hard evaluation case. This represents a second view of the observation presented in the terminal power analysis: the Lai procedure and our \OurMethod~each significantly outperform the SAVI procedure in the small-gap regime. 

Conversely, in \cref{fig:cumulativepower} (center), we illustrate a difficult setting in which the Lai procedure struggles. In this setting the performance gap is again 10 percentage points, but the distribution is lower-variance and skewed as compared to that of \cref{fig:cumulativepower} (right); as such, the Lai procedure cannot effectively adapt and our \OurMethod~significantly outperforms in terms of deciding more quickly for the alternative. 

\subsection{How sample efficient is STEP in practical policy comparison settings?}

In addition to the numerical validation, we evaluate \OurMethod~through two sets of real-world robot evaluation experiments 
 and a simulation experiment. In the first set of experiments, we compared policies with noticeable performance gaps to demonstrate the early-stopping capability of our approach. In the second set, we compared closely-competing policies to characterize necessary sample sizes for statistical validity when policy performance gaps become small. 
In the simulation experiment, we perform multi-task comparison of two SOTA imitation learning policies.
See \cref{subsec:task_details} in the Supplement for more details on the hardware experiments.

\subsubsection{Hardware Evaluation in the Large/Medium-Gap Regime}
In this experiment, we consider two manipulation tasks for a bimanual Franka Emika Panda robot: \textbf{FoldRedTowel} (\cref{fig:towel}) and \textbf{CleanUpSpill} (\cref{fig:spill_id} and \cref{fig:spill_ood}). We trained single-task diffusion policies~\cite{chi2023diffusion} on each task, with 300 human demonstrations for \textbf{FoldRedTowel} and 150 for \textbf{CleanUpSpill}, respectively. In addition to the RGB images, the policy receives the proprioceptive states as additional observations. 

In \textbf{FoldRedTowel}, we compare two policy checkpoints from a single training run. The baseline policy $\pi_0$ was trained for 10000 gradient steps with an AdamW~\cite{loshchilov2017decoupled} optimizer and achieved the action mean-squared error (MSE) of 1.61E-3 on the validation set. The other policy $\pi_1$ continued the training for 70000 additional steps, yielding the validation action MSE of 1.35E-3. To evaluate each policy, five in-distribution (ID) initial conditions were chosen and repeated 10 times each, constituting 50 total trials. ID initial conditions visually resemble the ones that appear in the training dataset. As shown in Table \ref{table:all_hardware_results} (rows 1-3), the empirical gap in success rates was 36 percentage points (56\% to 92\% success), suggesting that $\pi_0$ was under-trained. Each sequential method detected a significant difference at level $\alpha^* = 0.05$ in 17 -- 23 trials (apiece). That is, had the algorithms been active during collection, the last 27 -- 33 rollouts per policy were unnecessary for confirming the improvement of the later checkpoint over the former. Additionally, note that the Lai method and our \OurMethod~sequential procedures were each tuned for an $N_\text{max}$ of 50 (row 1), 200 (row 2), and 500 (row 3) rollouts. In the latter two cases, additional rollouts could have been run up to 200 or 500 per policy if the gap was smaller \emph{without compromising the validity of the decision}.

\begin{table*}[t!]
    \centering
    \resizebox{\linewidth}{!}{%
    \begin{tabular}{l|c|ccc|cc|ccc|c}
    \toprule
    Task & Type & $\alpha^*$ & $N_{\text{max}}$ & $N$ &  $\hat{p}_0$ & $\hat{p}_1$ & SAVI & Lai & STEP (Ours) & SPRT$^{***}$ \\
    \midrule
    FoldRedTowel &  $\pi_i$ & 0.05 & 50 & 50 & 0.560 & 0.920 & \textbf{20} & \textbf{17} & \textbf{19} & 17 \\
    FoldRedTowel &  $\pi_i$ & 0.05 & 200 & 50 & 0.560 & 0.920 & \textbf{20} & \textbf{21} & \textbf{21} & 17 \\
    FoldRedTowel &  $\pi_i$ & 0.05 & \underline{500} & 50 & 0.560 & 0.920 & \textbf{20} & \textbf{23} & \textbf{23} & 17 \\
    CleanUpSpill &  $\mathcal{D}^{i}_{s_0, o_0}$ & 0.05 & 50 & 50 & 0.280 & 0.800 & \textbf{7} & \textbf{8} & \textbf{8} & 7  \\
    CleanUpSpill &  $\mathcal{D}^{i}_{s_0, o_0}$ & 0.05 & 200 & 50 & 0.280 & 0.800 & \textbf{7} & \textbf{13} & \textbf{9} & 7  \\
    CleanUpSpill &  $\mathcal{D}^{i}_{s_0, o_0}$ & 0.05 & \underline{500} & 50 & 0.280 & 0.800 & \textbf{7} & \textbf{14} & \textbf{13} & 7  \\
    CarrotOnPlate & $\mathcal{D}^{i}_{s_0, o_0}$ & 0.05 & 100 & 100 & 0.590 & 0.680 & \text{--} & \text{--} & \text{--} & \text{--} \\
    CarrotOnPlate & $\pi_i$ & 0.05 & 100 & 100 & 0.680 & 0.760 & \text{--} & \text{--} & \text{--} & \text{--}  \\
    \midrule
    SpoonOnTowel & $\pi_i$ &  0.01 & 500 & 500 & 0.084 & 0.386 & \textbf{33} & \textbf{36} & 
    \textbf{36} & 26 \\
    EggplantInBasket & $\pi_i$ & 0.01  & 500 & 500 & 0.400 & 0.564 &  192 & \textbf{125} & \textbf{131} & 128 \\
    StackCube & $\pi_i$ & 0.01  & 500 & 500 & 0.000 & 0.030 &  329 & 417 & 225 & 135  \\
    \midrule
    Multitask & $\pi_i$ & 0.03 & 1500 & 1500 & \text{N/A} & \text{N/A} & 554 & 578 & 392 & 289\\
    \bottomrule
    \end{tabular}
    }
    \caption{Empirical time-to-correct-decision for all \textbf{hardware (top)} and \textbf{simulation (bottom)} policy comparisons. The comparison type is described first; $\pi_i$ is comparing two policies, $\mathcal{D}^{i}_{s_0, o_0}$ compares one policy under possible distribution shift. The utilized Type-I Error $\alpha^*$ and $N_\text{max}$ describe the constraints applied \textit{a priori} by the evaluator (we \underline{underline} to emphasize the change in $N_\text{max}$ for rows 1-3 and 4-6; observe that the sensitivity of the stopping times is very small). $N \leq N_\text{max}$ represents the amount of data available for the statistical analysis. We report the terminal empirical success rates (after $N$ trials) of each policy in each setting under $\hat{p}_i$ (this information is not available to any feasible algorithm). We do not have truth labels on this data; however, in all cases, every method arrived at the same decision, including the Oracle SPRT which has \textit{a priori} access to privileged information $(\hat{p}_0, \hat{p}_1)_N$; this decision was Reject Null for all rows except the CarrotOnPlate tasks, which each returned Fail To Decide. We report the stopping times of all methods on the right of the table for every context; in all cases: lower is better. We put in \textbf{bold} any \emph{feasible} method result that is near-optimal within ten trials (absolute) or 25\% (relative) of the SPRT Oracle, which is \emph{not implementable} by an evaluator. In the Multitask setting, we test $p_1 > p_0$ \textit{uniformly across the preceding three tasks}. This stopping time is the sum by column of the stopping times for the three tasks. Our method saves the evaluator over 160 trials in uniform certification over these three tasks as compared to either feasible baseline.}
    \label{table:all_hardware_results}
\end{table*}

In \textbf{CleanUpSpill}, we compare the same policy on two different sets of initial conditions. The task is similar to the one originally presented by~\citet{xu2025can}, which compares a set of ID initial conditions against the out-of-distribution (OOD) initial conditions. The ID set includes 10 initial conditions with a white towel and a short blue mug whereas the OOD set uses 10 with a checkered towel with a tall cyan mug (each initial condition is repeated five times). As shown in \cref{table:all_hardware_results} (rows 4--6), the empirical gap of 52 percentage points was detected in 7--14 trials by all methods, though they were tuned (where applicable) to an $N_\text{max}$ up to thirty to seventy times larger; this demonstrates the significant reduction in sensitivity (from an evaluator's standpoint) arising from setting $N_\text{max}$ versus choosing a batch size $N$. Furthermore, \OurMethod's efficiency only minimally degrades when $N_\text{max}$ is increased from 200 (row 5) to 500 (row 6). In this setting, any of the three sequential methods would have prevented the need for at least 70 of the 100 total rollouts (35 of the 50 batch trials per policy). 

\subsubsection{Hardware Evaluation in the Small-Gap Regime} 
\begin{figure*}[!thb]
    \centering
    \begin{minipage}[t]{0.3\linewidth}
        \centering
        \includegraphics[width=\textwidth, trim=0 0 0 0, clip]{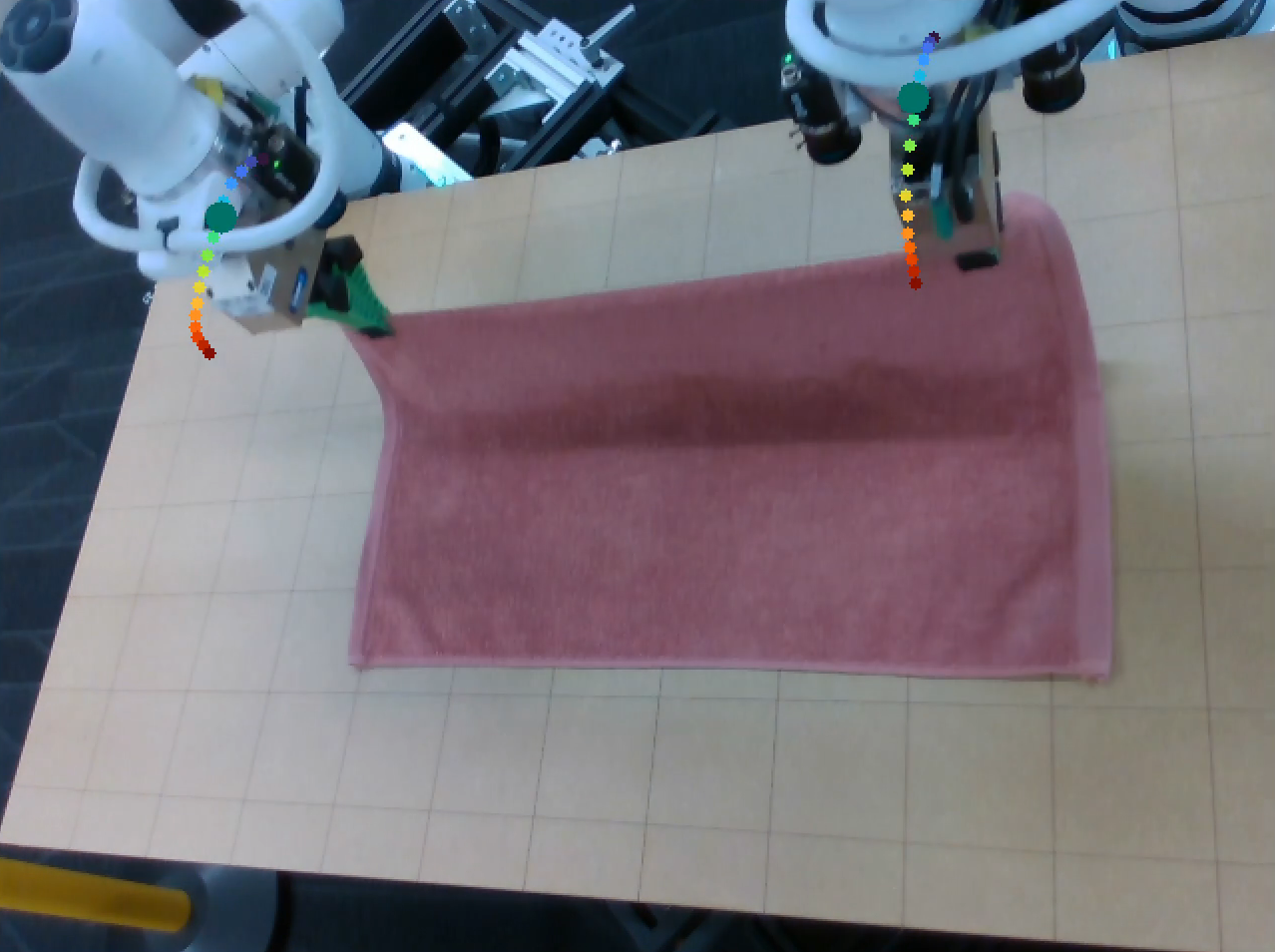}
        \subcaption{\textbf{BimanualFoldRedTowel}}
        \label{fig:towel}
    \end{minipage}
   \begin{minipage}[t]{0.3\linewidth}
        \centering
        \includegraphics[width=\textwidth, trim=0 0 0 0, clip]{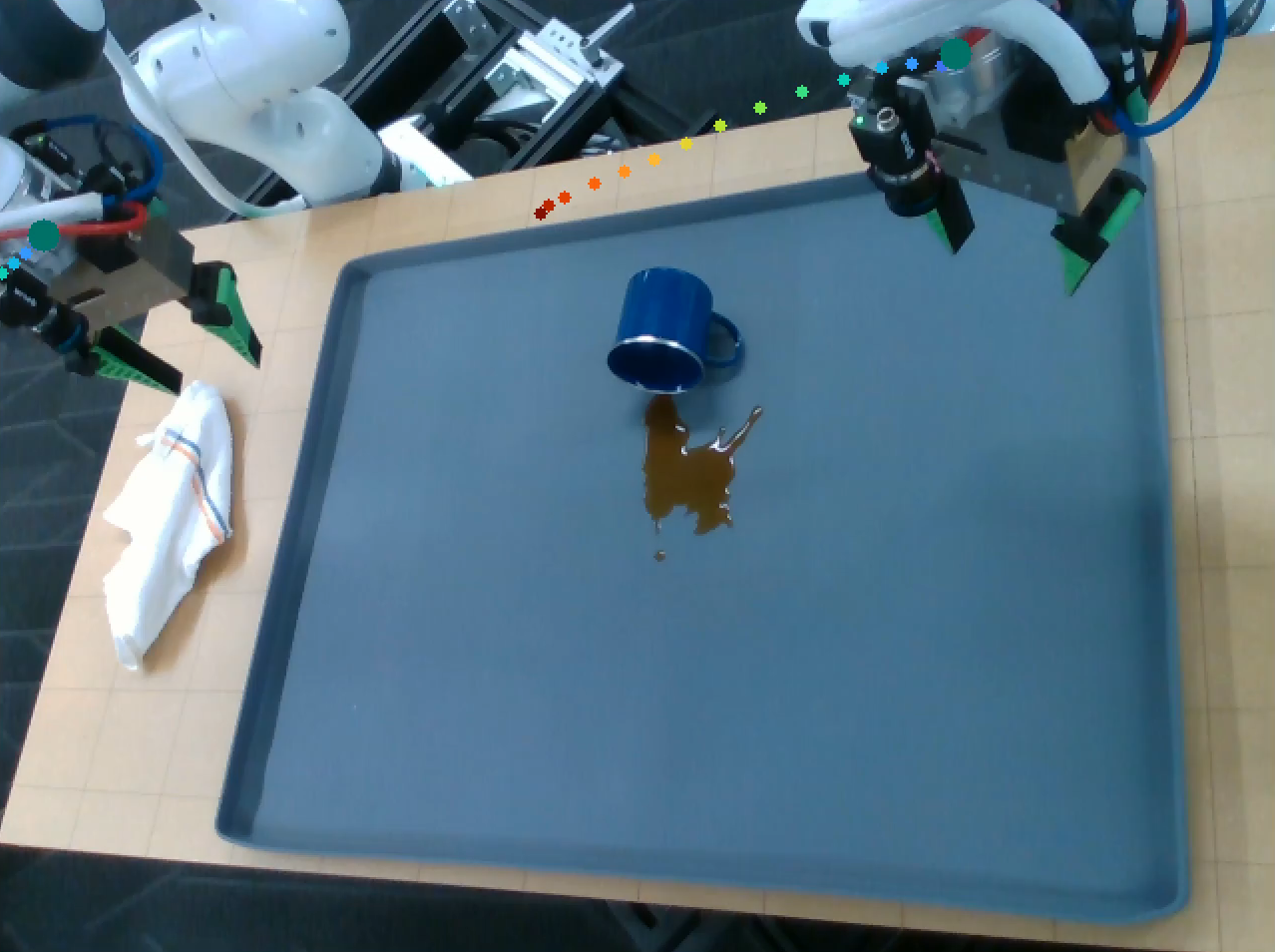}
        \subcaption{\textbf{BimanualCleanUpSpill} (ID)}
        \label{fig:spill_id}
    \end{minipage}
    \begin{minipage}[t]{0.3\linewidth}
        \centering
        \includegraphics[width=\textwidth, trim=0 0 0 0, clip]{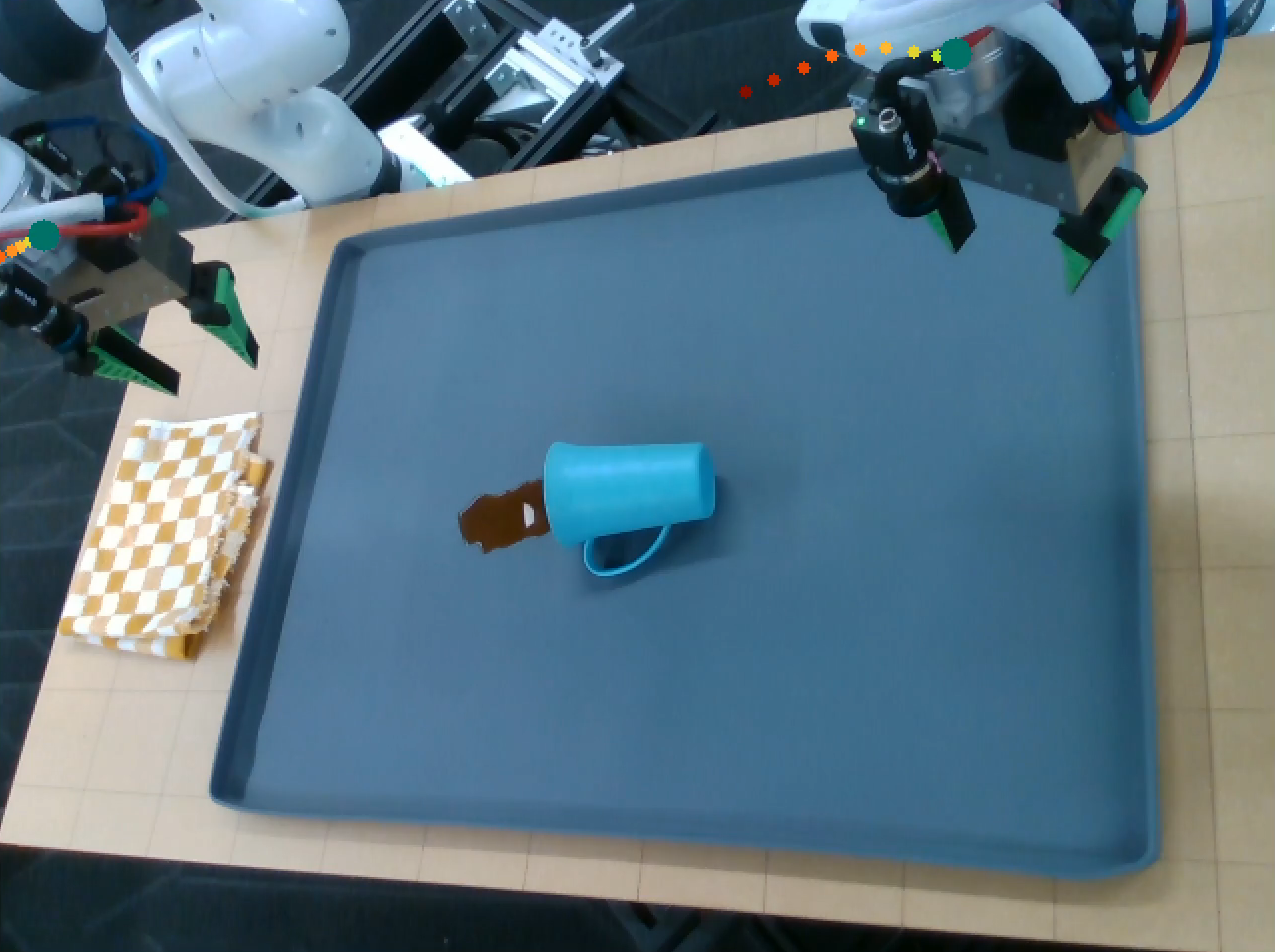}
        \subcaption{\textbf{BimanualCleanUpSpill} (OOD)}
        \label{fig:spill_ood}
        \vspace{3ex}
    \end{minipage}
    \begin{minipage}[t]{0.3\linewidth}
        \centering
        \includegraphics[width=\textwidth, trim=0 0 0 0, clip]{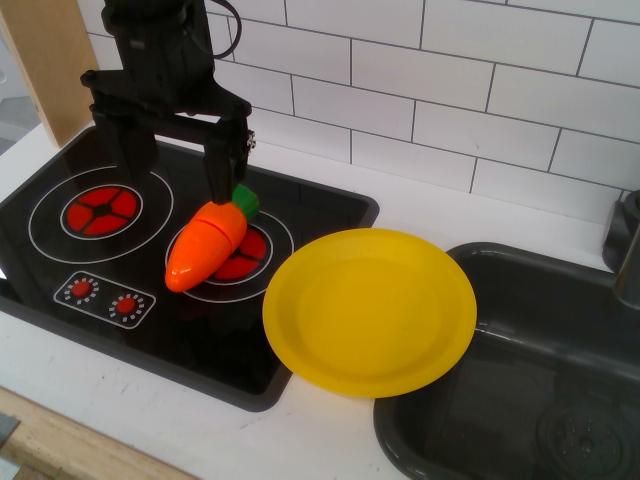}
        \subcaption{\textbf{PutCarrotOnPlate} (no distractors)}
        \label{fig:no_dist}
    \end{minipage}
    \begin{minipage}[t]{0.3\linewidth}
        \centering
        \includegraphics[width=\textwidth, trim=0 0 0 0, clip]{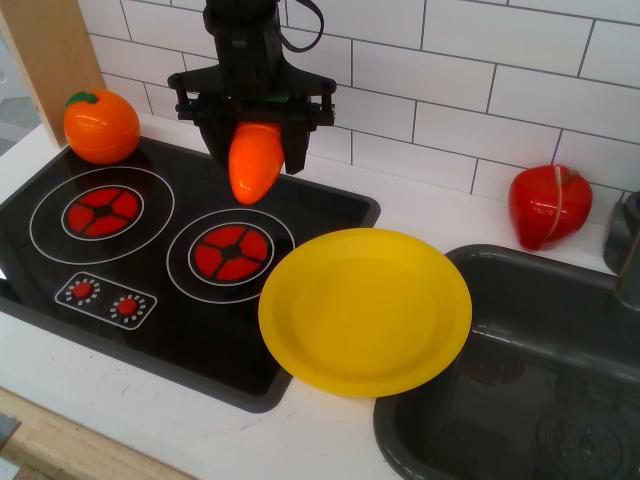}
        \subcaption{\textbf{PutCarrotOnPlate} (with distractors)}
        \label{fig:with_dist}
    \end{minipage}
    \caption{Snapshots of robot policy evaluation tasks. (Top) Bimanual manipulation tasks with diffusion policy. Colored dots represent the camera projection of planned future end-effector positions. In \textbf{BimanualFoldRedTowel}, all the evaluations are done with in-distribution (ID) initial conditions and we compare two policy checkpoints from a single training run. In \textbf{BimanualCleanUpSpill}, we evaluate a single policy checkpoint in ID initial conditions with a white towel and out-of-distribution (OOD) initial conditions with a green towel to measure generalization performance. (Bottom) \textbf{PutCarrotOnPlate} task on the WidowX platform in a toy kitchen environment. The carrot is initially placed in one of three possible locations on the stove. The environment can either have no distractors or two distractors. We compare Octo and OpenVLA under the nominal environment distribution, and compare Octo performance in nominal environment distribution and under distribution shift. Detailed policy comparison metrics are given in Table~\ref{table:all_hardware_results}. }
    \label{fig:task_visuals}
\end{figure*}

We evaluate \OurMethod~on two open-source vision-language-action (VLA) models: Octo-Base \cite{octo_2023}, an action-chunking transformer-based diffusion policy, and OpenVLA \cite{kim2024openvla}, an autoregressive policy leveraging a pretrained large language model backbone. All experiments were conducted in a toy kitchen environment from the Bridge Data V2 dataset \cite{walke2023bridgedata}, which is included in both policies' training data. We considered the task of placing a carrot on a plate (denoted \textbf{CarrotOnPlate}, see \cref{fig:no_dist} and \cref{fig:with_dist}), which is representative of evaluations investigated in \cite{octo_2023,kim2024openvla}. All policies were run on the Widow X 250S following the setup in \cite{walke2023bridgedata}.

For this task, the initial placement of the carrot is uniformly sampled from three possible locations (left, center, or right) on the counter, and the plate is placed next to the sink (see Figure~\ref{fig:no_dist}). At the start of each trial, the robot joint angles are initialized such that the gripper is roughly aligned with the carrot initial position. The environment follows a categorical distribution with two outcomes: no object distractors or two object distractors. We consider two environment distributions: \textbf{Env1} in which there are no distractors with probability \(0.8\) and \textbf{Env2} in which there are no distractors with probability \(0.6\). Object distractors are sampled uniformly (without replacement) from the following object categories: orange, apple, green and blue sponges, brown and yellow cubes, eggplant, spoon, and towel. The initial locations of distractors is also sampled uniformly without replacement from four possibilities: on the stove, left of the stove, above the stove, or next to the faucet. The distractors are physically placed according to a uniform (continuous) distribution within the selected region. We sample 100 environment configurations each for three settings: i) Octo under distribution \textbf{Env1}, ii) OpenVLA under distribution \textbf{Env1}, and iii) Octo under distribution \textbf{Env2}. We then make the following comparisons with~\OurMethod: i) Octo (\textbf{Env1}) and Octo (\textbf{Env2}), and ii) Octo (\textbf{Env1}) and OpenVLA (\textbf{Env1}). In the first comparison (Table \ref{table:all_hardware_results}, row 7), we test the effect of distribution shift in the probability of distractors being present. Here, the $\hat{p}_0$ corresponds to the perturbed distribution Octo (\textbf{Env2}), and $\hat{p}_1$ to the nominal distribution Octo (\textbf{Env1}). We find that, while there is an empirical gap (59\% vs 68\%) at $N = 100$, no method returns a significant result at $\alpha^* = 0.05$. In the second comparison (Table \ref{table:all_hardware_results}, row 8), $\hat{p}_0$ corresponds to Octo (\textbf{Env1}) and $\hat{p}_1$ corresponds to OpenVLA (\textbf{Env1}). We observe no significant gap despite the empirical gap of 8 percentage points in favor of OpenVLA. 
Importantly, insignificance of the test does not mean that the null hypothesis should be accepted~\cite{greenland2016statistical}; there remains a possibility that OpenVLA actually outperforms Octo, or that Octo's performance indeed degrades due to the presence of distractors. However, our budget of $N_{\max} = 100$ was likely not sufficient to accumulate enough evidence.
In \cref{subsec:carrot_analysis} in the Supplement, we further investigate this data insufficiency to show that, if the ground truth values were equal to the empirical success rates (59\% vs. 68\% and 68\% vs. 76\%), then we would require $N_{\max} = 500$ trials to confidently determine $p_1 > p_0$. This number is an order of magnitude larger than the current norms, reflecting fundamental yet often overlooked challenges in trustworthy policy comparison.

\subsubsection{Multi-Task Evaluation in SimplerEnv Simulation}
\label{subsubsec:Multitask}
Finally, we briefly consider the problem of multi-task and multi-policy extensions to this framework, and illustrate its potential via an example of policy evaluation in simulation (where costs are lower than on hardware, but still can be significant). Concretely, Octo-Small ($\pi_1$) and Octo-Base ($\pi_0$) \cite{octo_2023} are compared in the SimplerEnv~\cite{li2024evaluating} simulation environment on three tasks of varying empirical difficulty. On the \textbf{EggplantInBasket} task, the policies each succeed at a near-50\% rate, with a gap of 16.4 percentage points. For the \textbf{SpoonOnTowel} task, the gap is larger at 30 percentage points. For the hardest task, \textbf{StackCube}, the performance gap is 3 percentage points. The evaluation statistics are shown in Table \ref{table:all_hardware_results}, rows 9-11. Note that the empirical success rates we observed are consistent with the findings of \citet{li2024evaluating} that Octo-Small is more performant on these tasks (see their Table V). We seek to evaluate the multitask comparison of the two policies. Letting $p_s^{[\tau]}$ denote the performance of Octo-Small and $p_b^{[\tau]}$ denote the performance of Octo-Base on task $\tau$, we test:
\begin{equation}
    \begin{split}
    \label{Eqn:MultiHypothesis}
        &\mathbb{H}_0: \exists \tau \in \{1, 2, 3\}~ p_s^{[\tau]} \leq p_b^{[\tau]} \\
        &\mathbb{H}_1: \forall \tau \in \{1, 2, 3\}~ p_s^{[\tau]} > p_b^{[\tau]}.
    \end{split}
\end{equation}
Many established and sophisticated methods exist to efficiently run multi-hypothesis testing (in this case, we are essentially evaluating three separate hypotheses, one for each task\footnote{Note that multi-hypothesis testing can naturally handle the case of \textbf{multi-policy} comparison as well, where we would reduce the test to a set of pairwise comparisons that are examined simultaneously.}). As a simple illustration, we use the standard Bonferroni (union bound) correction \cite{Dunn_1961} to evaluate the test: specifically, running each of the individual three tests at level $\alpha^*_{[\tau]} = 0.01$, we observe the stopping times shown in Table \ref{table:all_hardware_results}, rows 9-11 (right-hand side). Via the Bonferroni correction, the combined decision (\textbf{RejectNull}, because every subtest decided \textbf{RejectNull}) expressed in \cref{Eqn:MultiHypothesis} is then confirmed at $\alpha = 0.03 = \sum_\tau \alpha^*_{[\tau]}$. As illustrated in \cref{table:all_hardware_results}, each sequential method saves a substantial number of simulation rollouts on the easiest subtest (\textbf{SpoonOnTowel}). As expected, SAVI begins to struggle when the tests become more challenging, such as in \textbf{EggplantInBasket}. Finally, we observe the weakness of the Lai procedure: in heavily skewed cases, it suffers substantially even compared with SAVI methods, as shown in \textbf{StackCube}. To summarize: naive multitask evaluation requires the aggregation of multiple batches of rollouts, here totaling 1500 per policy (500 per task per policy). Note in rows 9-11 how different the number of requisite trials can be, and therefore how hard it is to reliably run the evaluation using a fixed batch size. On the easiest task, even when tuned to $N_\text{max} = 500$, the comparison was answered in fewer than 40 rollouts by all sequential methods, a savings of over 90\%. On the progressively harder cases the number of required samples increased 5-10 times over the easiest, but our method (\OurMethod) improved substantially over each of the other sequential procedures. In total, \OurMethod~would have saved the evaluator an additional 160 rollouts for each of Octo-small and Octo-base for the multitask comparison problem as compared to the current SOTA approaches.

\section{Related Work}
\label{sec:related_work}
Thus far, our discussion has focused solely on the practical setting of comparative policy evaluation. However, mathematical statistical testing has a well-established near-century-long history, and sequential testing in particular has existed for nearly the same amount of time. This section provides an extensive review of the statistics literature to further highlight the significance of our approach.

\subsection{Statistical Testing and Policy Evaluation}

The Neyman-Pearson (N-P) statistical testing paradigm~\cite{neyman_1933} forms the foundation of the last century of frequentist statistical decision theory. 
Applicable to questions spanning many fields, these methods have generally been applied in robotics for predictable policy \emph{characterization}\footnote{As a simple example, one can accurately predict \textit{a priori} that for estimating Ber($p$) with $\hat{p} \in [0.25, 0.75]$ and $N \geq 36$, a 95\% confidence interval for $p$ will be approximately $\hat{p} \pm \frac{1}{\sqrt{N}}$.}~\cite{Vincent_2024, agarwal2021deep}, where the number of samples is fixed and a single decision or estimate is to be made. The Neyman-Pearson Lemma \cite{neyman_1933} and the Karlin-Rubin Theorem \cite{karlin_theory_1956} give sufficient conditions for maximal power probability ratio tests (PRTs), which form the precursor to the SPRT Oracle used in this work. 
Specific methods have been developed for robot policy comparison-type problems in the context of 2x2 contingency tables. Of the tests by \citet{Fisher_1922}, \citet{boschloo_1970}, and \citet{barnard_significance_1947}, the latter is a powerful batch procedure for comparison problems. However, while it has strong power in the batch setting, it does not provide a direct mechanism for choosing the appropriate size of the batch \textit{a priori}. 

\subsection{Sequential Statistical Evaluation Methods}
The difficulty in choosing the appropriate batch size motivates the sequential testing framework set out in \citet{wald_sequential_1945}. This is the setting adopted in this paper and described in \cref{sec:problem}. \citet{wald_optimum_1948} showed that in the simple-vs-simple setting, the sequential probability ratio test (SPRT) minimizes the expected number of samples among all tests that control Type-I and Type-II error, extending the N-P result. Significant efforts have extended near-optimality results to the composite testing regime. Minimax results attempt to limit the worst-case expected sample size~\cite{kiefer_properties_1957, lorden_nearly-optimal_1977, fauss_minimax_2020}, while others minimize the expected sample size under a weighted mixture over the alternatives~\cite{Schwarz_1962, Fortus_1979}. \citet{lai_nearly_1988} reconciled this Bayesian interpretation with the frequentist developments of Chernoff~\cite{chernoff_sequential_1961, chernoff_sequential_1965_III, chernoff_sequential_1965_IV}.

\subsubsection{Optimal Stopping-Based Methods}
The direct approach to synthesizing near-optimal decision regions in the composite-vs-composite regime relies on developments in the theory of martingales and optional stopping~\cite{Williams_1991}. \citet{van_moerbeke_optimal_1974} provides a clear reduction of the statistical decision making problem to that of optimal stopping and exposits previous developments in the area equating the optimal stopping procedure with the solution of a Stefan-type free-boundary partial differential equation (PDE)~\cite{caffarelli_geometric_2005}. Unfortunately, while specification of the testing problem is usually feasible, its mapping to the PDE parameterization is generally difficult to specify under composite null hypotheses, rendering this method impractical. A parallel line of work considers utilizing asymptotic approximations of the solution to the free-boundary problem in order to construct near-optimal tests. The work of \citet{lai_nearly_1988} solves for a near-optimal procedure in the univariate composite-vs-composite setting and follow-on work~\cite{lai_nearly_1994, chan_asymptotic_2000} extends this to the multivariate setting. While useful for proving asymptotic optimal rates, these methods suffer in the finite-$N_{\max}$ regime due to losing decision-making information. 

\subsubsection{Safe, Anytime-Valid Inference (SAVI) Methods}
The difficulty in synthesizing optimal procedures in the multivariate setting (see, e.g., \cite{lai_nearly_1994, chan_asymptotic_2000}) motivates recent developments which fuse a distributionally robust safety invariance and the structure of the probability ratio test. This yields a family (SAVI) of methods which generalize to any $N_{\max}$ (i.e., arbitrarily small performance gaps) and a richer set of composite-hypothesis testing problems. Utilizing Ville's Inequality (a sequential generalization of Markov's Inequality)~\cite{Ville_1939}, SAVI methods construct a probability ratio test that enforces Type-I error control uniformly in time~\cite{howard_time-uniform_2021, ramdas_game-theoretic_2023}. In the small sample regime it is the effect of the Power-1 nature of the test~\cite{Robbins_1974, Lai_1977_PowerOne, ramdas_game-theoretic_2023} that can most negatively affect performance.

\subsection{Numerical Implementations}
Numerical methods are of significant practical importance in statistical testing. Excellent implementations of batch procedures have been released in the SciPy~\cite{2020SciPy-NMeth} package; recently, the particular problem of optimal binomial confidence intervals was addressed in~\cite{Vincent_2024}. However, despite a significant history dating back to the 1980s~\cite{Chernoff_1986}, numerical methods for sequential analysis in evaluation are relatively limited. Further, in~\cite{Chernoff_1986} as in more recent work, emphasis has remained on the simple-null or univariate composite setting~\cite{Novikov_2022, Novikov_2023, fauss_minimax_2020} for non-SAVI procedures. However, to our knowledge, a numerical implementation of near-optimal policy comparison procedures in the full multidimensional composite-vs-composite setting has not yet been implemented.

\section{Limitations}
\label{sec:Limitations}
Despite the strong performance of \OurMethod~as shown in \cref{sec:experiments}, it does have several practical and theoretical limitations. First, the development of \OurMethod~heavily relies on the mathematical structure of Bernoulli distributions, which essentially requires the policy evaluation results to be presented as binary success/failure metrics. Extending \OurMethod~to more complex discrete and continuous distributions would be a valuable future research direction for broader applicability. Second, in our current implementation a user needs to specify \OurMethod's risk accumulation rate $f(k)$. Although we empirically show that the uniform rate already achieves near-optimality, further research is needed to improve the performance. 

Besides these limitations that are specific to \OurMethod, researchers must be mindful of underlying assumptions and common misuse of statistical hypothesis testing~\cite{greenland2016statistical}.
Concretely, the foundational assumption is that evaluation data are i.i.d. as all the statistical assurances are built on top of it. To this end, we ensured that all hardware evaluations comported with the best practices of \cite{kress-gazit_robot_2024}. Finally, Goodhart's Law \cite{Strathern_1997} provides a useful warning about inadvertent p-hacking: having statistically rigorous evaluation is important, but if ``significance at level $\alpha^*$" becomes a target and not a metric, it can induce undesirable research practices. We emphasize that p-values and significance levels are only as rigorous as the rigor of the research methods that utilize them. Their adoption has the potential to be enormously valuable for the empirical codification of robotics knowledge, but they are not a panacea. 

\section{Future Work} 
\label{sec:futurework}
There are several concrete extensions of STEP which promise significant benefits to practical robotic evaluation. Most direct is incorporation of more principled multi-policy evaluation, building upon \cref{subsubsec:Multitask}. Additionally, partial credit schema have found increasing importance in long-horizon evaluation~\cite{kress-gazit_robot_2024}; generalizing STEP beyond binary success/failure metrics to discrete (multinoulli) partial credit is direct in theory, although it presents additional challenges in implementation. 

\section{Conclusion} 
\label{sec:conclusion}
We present \OurMethod, a novel sequential statistical method to rigorously compare performance of imitation learning policies through a series of evaluation trials. \OurMethod's sequential evaluation scheme provides flexibility in adapting the number of necessary trials to the underlying difficulty of policy comparison. This leads to sample efficiency in cases where one policy clearly outperforms the other, while avoiding overconfident and potentially incorrect evaluation decisions when the policies are closely competing. 
We show that STEP near-optimally minimizes the expected number of trials required in the policy comparison problem. Furthermore, \OurMethod~matches or exceeds the performance of state-of-the-art baselines across a wide swath of practical evaluation scenarios in numerical and robotic simulation and on numerous physical hardware demonstrations.
These results highlight the practical utility of \OurMethod~as a versatile statistical analysis tool for policy comparison, contributing to the foundation of robot learning as an empirical science.

\section*{Acknowledgments}
This work was primarily completed during an internship at Toyota Research Institute (TRI). Anirudha Majumdar was partially supported by the NSF CAREER Award [\#2044149] and the Sloan Fellowship. The authors would like to thank Chen Xu for insightful feedback and discussions, as well as the robot teacher team at TRI headquarters for their data collection efforts, in particular Derick Seale and Donovan Jackson. Finally, we thank the anonymous reviewers and the area chair for incisive suggestions for improving the paper. 

\clearpage

\bibliographystyle{plainnat}
\bibliography{main_camera}

\newpage{}

\section{Supplement}
\label{sec:Supplement}

\begin{figure*}[htbp!]
    \centering
    \includegraphics[width=\textwidth] {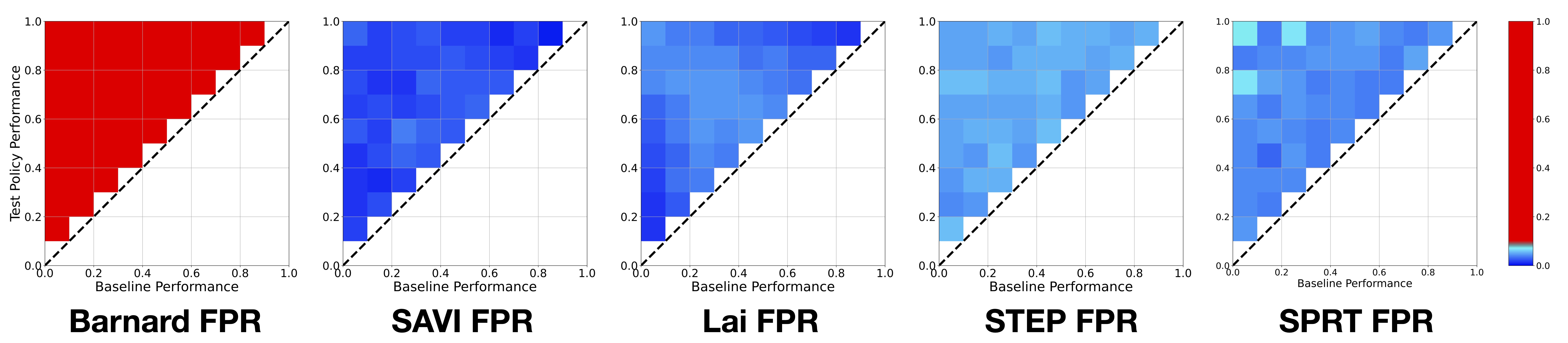} 
    \caption{False positive rate of four feasible methods (Barnard, SAVI, Lai, and Ours (\OurMethod)) and the SPRT (Oracle method) for 1000 simulated trajectories on each of 45 alternatives (squares in color); $N_\text{max}=100$ and $\alpha^* = 0.05$. Note that naively utilizing a batch method in sequence leads to violation of Type-1 Error control (Barnard). Additionally, note that SAVI and Lai struggle to utilize the full risk budget in finite $N_\text{max}$ (darker blue regions). }
    \label{fig:terminal_fpr_N100}
    \vspace{1ex}
\end{figure*}
\begin{figure*}[htbp!]
    \centering
    \includegraphics[width=\textwidth] {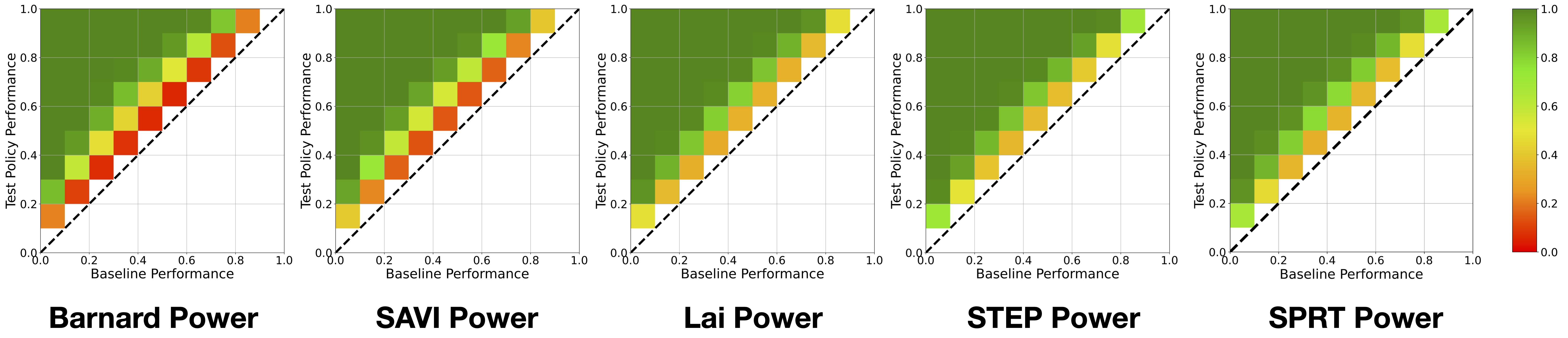} 
    \caption{Terminal power of four feasible methods (Barnard, SAVI, Lai, and Ours (\OurMethod)) and the SPRT (Oracle method) for 5000 simulated trajectories on each of 45 alternatives (squares in color); $N_\text{max}=100$ and $\alpha^* = 0.05$. Because $N_\text{max}$ is small, the terminal power is generally low for gaps less than 20 percentage points. Moving from left to right: sequentializing Barnard's Test is inefficient due to a loss of structure; SAVI methods also suffer when $p_0$ and $p_1$ are closely competing, due to the method inherently generalizing to arbitrary $N_\text{max}$. The Lai procedure and our \OurMethod~are similar to the SPRT oracle; however, note that Lai struggles more at the extremes (bottom left and top right). This inefficiency in the skewed regime becomes more pronounced as $N$ grows and the gaps shrink.}
    \vspace{1ex}
    \label{fig:terminalpower_N100}
\end{figure*}

\begin{figure*}[!thb]
    \centering
    \includegraphics[width=\textwidth] {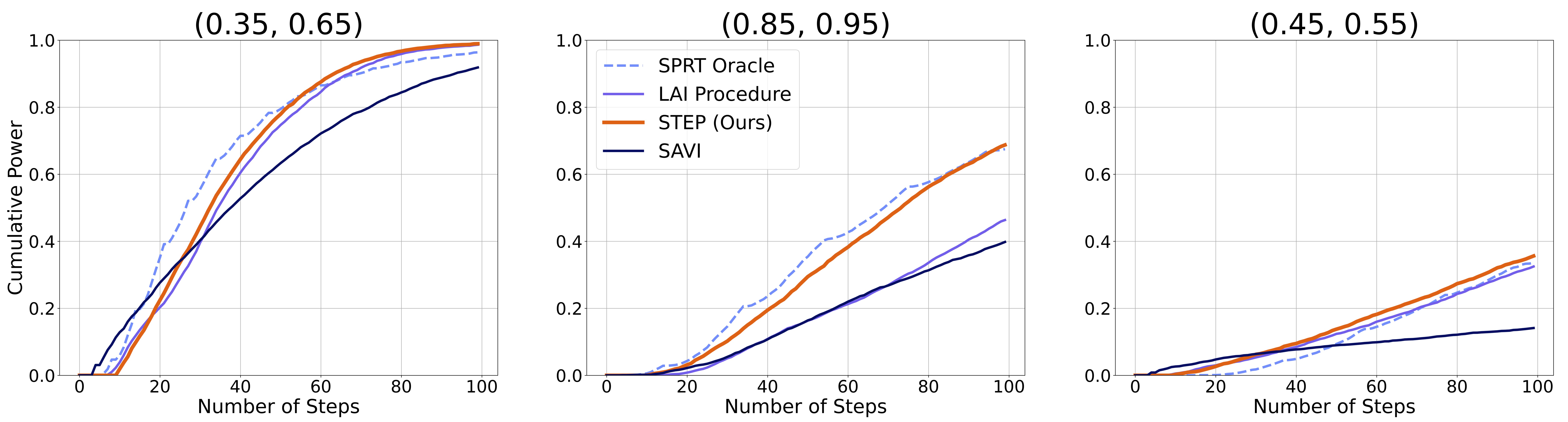}
    \caption{Cumulative power of all feasible methods (Lai, SAVI, STEP (Ours)) and SPRT Oracle over 5000 trajectories in three evaluation settings of increasing difficulty; $(p_0, p_1)$ for each setting title the respective figures. $N_\text{max}=100$ and $\alpha^* = 0.05$. The expected time-to-decision is the integral of the area \emph{above} the cumulative power curve; therefore, curves higher and to the left are better. (Left) For a gap of 30 percentage points, all methods demonstrate similar stopping times. (Center) For a gap of 10 percentage points in the low-variance regime (i.e., farther from 0.5), STEP significantly outperforms the Lai and SAVI procedures. (Right) For a gap of 10 percentage points in the high-variance regime, STEP and Lai are similar but again SAVI struggles and underperforms the other methods. }
    \label{fig:cumulativepower_N100}
    \vspace{1ex}
\end{figure*}

\subsection{Details of Real-World Robot Experiments}
\label{subsec:task_details}
All of our real-world hardware tasks are visualized in \cref{fig:task_visuals}.
In \textbf{FoldRedTowel}, the robot first observes an unfolded red towel placed in random poses. The task is considered a success if the robot folds the towel twice and then moves the folded towel to a corner of the table. In \textbf{CleanUpSpill}, a mug is initially lying sideways on the table and a coffee spill exists near the mug. The task is successful if one arm puts the mug upright while the other arm picks up a white towel and wipes the spill. In both tasks, a total of four RGB cameras observe the Franka robot and the objects, where two monocular cameras are mounted on the table top and a stereo wrist camera on each of the arms. We trained single-task diffusion policies~\cite{chi2023diffusion} on each task, with 300 human demonstrations for \textbf{FoldRedTowel} and 150 for \textbf{CleanUpSpill}, respectively. In addition to the RGB images, the policy receives the proprioceptive states as additional observations. Following \citep{chi2023diffusion}, the image observations are passed to the ResNet-18~\cite{he2016deep} encoder before fed into the U-Net~\cite{ronneberger2015u} diffusion policy architecture. $T_o = 2$ observations are stacked and fed into the policy network to predict $T_p = 16$ steps of actions. The actions are re-planned after $T_a = 8$ actions are executed.

For the \textbf{CarrotOnPlate} task, an experiment is recorded as a success if the robot policy succeeds in placing the carrot on the plate within the max episode count without: i) pushing the carrot off the counter, ii) colliding with the back wall, iii) pushing the plate into the sink, and iv) accumulating a total of 3 cm of negative \(z\) commands when the end-effector is in contact with the table surface. For Octo evaluations, we use an action-chunking horizon of 2. 

In all the experiments, we take the effort to mitigate distribution shift during trials, such as a change in lighting conditions. We also randomize the order of trials so that any distribution shift due to other factors (e.g., hardware degradation over the course of trials) is equally reflected in all the settings. Where applicable, we also separate the role of the evaluator from the demonstrator of the tasks for training. These practices are adopted from \cite{kress-gazit_robot_2024} to reduce unintended variability in environmental conditions during policy evaluation.

\subsection{Additional Numerical Simulation Results}
\label{subsec:results_Nmax_100}
We plot results analogous to \cref{fig:terminalpower}, \cref{fig:terminal_fpr_N500}, and \cref{fig:cumulativepower} in \cref{sec:experiments} for $N_\text{max} = 100$ and $\alpha^*=0.05$ in \cref{fig:terminalpower_N100},  \cref{fig:terminal_fpr_N100}, and \cref{fig:cumulativepower_N100}. We include for this case the power profile for a Barnard Test that is validly sequentialized (using Bonferroni's correction); \emph{this rectifies the Type-1 Error violation in \cref{fig:terminal_fpr_N100}}. In so doing, it loses significant power and fails to meaningfully compete with the SOTA sequential procedures. In additional to inefficient computational properties, the Bonferroni-correct Barnard procedure becomes even weaker for larger $N_\text{max}$.

A key point of emphasis in the $N_\text{max}=100$ regime is the low power of all tests for gaps of approximately 10 percentage points and smaller. Notably, no procedure has power over 50\% in the hardest regimes (see the orange regions of every method in \cref{fig:terminalpower_N100}). A small amount of this is due to the sequential procedure; however, a significant amount reflects fundamental uncertainty (variance in outcomes) present for small sample sizes in evaluation. The implication of this is the need for significant increases in evaluation trials in order to effect meaningful comparisons when the underlying gap is small. This will be considered further in the context of the \textbf{CarrotOnPlate} hardware experiments (\cref{sec:experiments}) in \cref{subsec:carrot_analysis} below.

Finally, we note the presence of a small hint to the weakness of the Lai procedure in skewed settings. Note that in the bottem left and top right of the Lai panel of \cref{fig:terminalpower_N100}, the power significantly lags STEP and SPRT; in a similar vein, note the regions of darker blue in the Lai procedure panel of \cref{fig:terminal_fpr_N100}. These reflect an inherent inefficiency undergirding Lai method, which directly explain the significant gap on the highly-skewed \textbf{StackCube} task in \cref{sec:experiments}. 

\subsection{Empirical Results with Regenerated Sequences}
\label{subsec:BigTableRegenerated}
To (approximately) evaluate the counterfactual noise in the data generation process for robotic evaluation, we randomly generate Bernoulli sequences using (as the true data-generating parameters) the empirical success rates of each task in \cref{table:all_hardware_results}. This provides an estimate of the average sample complexity for each method \emph{were the empirical success rates equal to the true rates `in the world'}, providing a Bernoulli counterfactual in \cref{table:averaged_bernoulli_counterfactual_hardware_results}; in that table, all entries present the empirical mean complexities (with standard deviation) over 400 regenerated sequences per task.
\subsection{Further Analysis of \textbf{CarrotOnPlate} Experiments}
\label{subsec:carrot_analysis}
\begin{figure*}[!thb]
\centering
\begin{minipage}[t]{0.32\linewidth}
    \centering
    \includegraphics[width=0.9
    \textwidth]{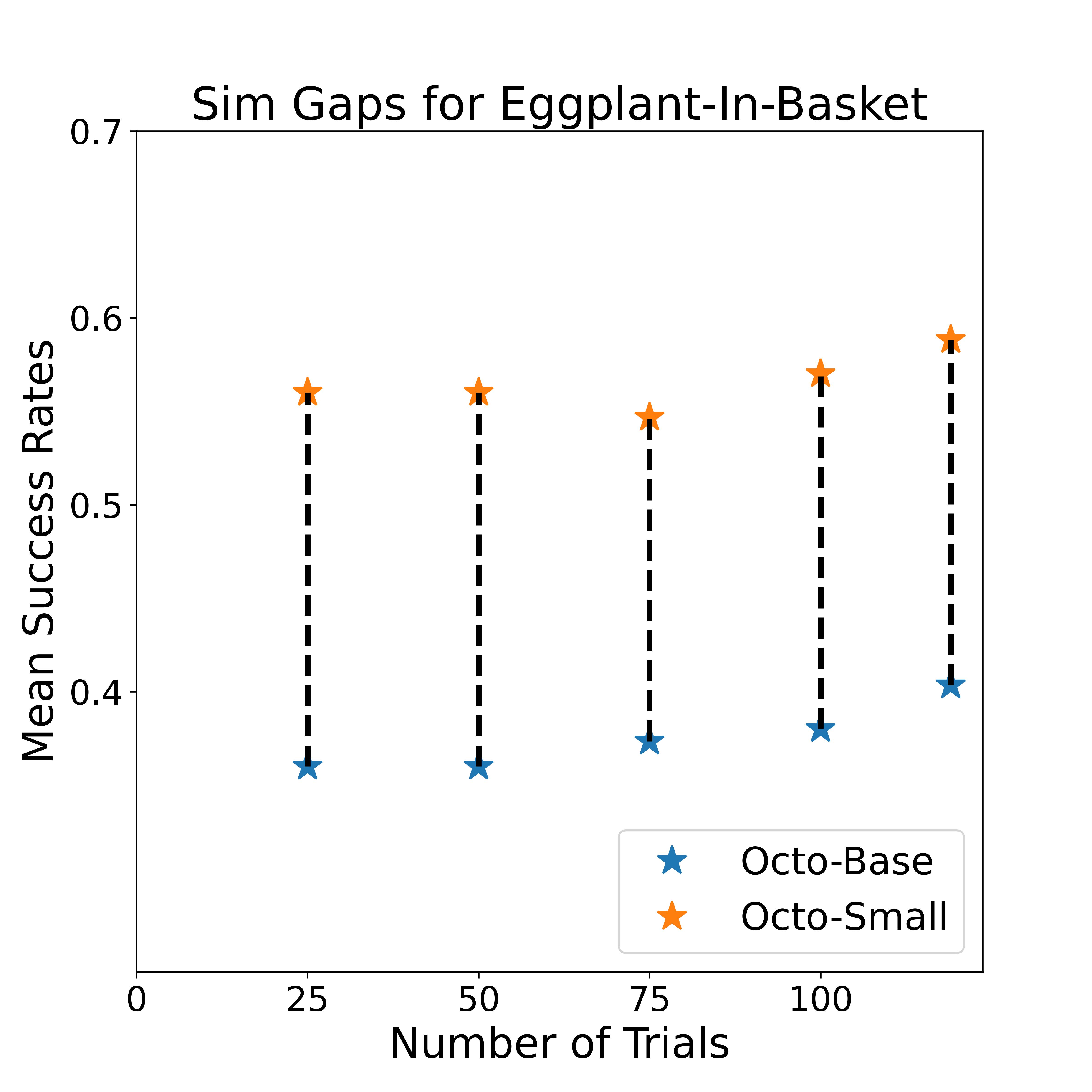}
    \subcaption{Octo-Base and Octo-Small in simulation \textbf{EggplantInBasket} task}
    \label{fig:EggplantInBasket}
\end{minipage}
\begin{minipage}[t]{0.32\linewidth}
    \centering
    \includegraphics[width=0.9
    \textwidth]{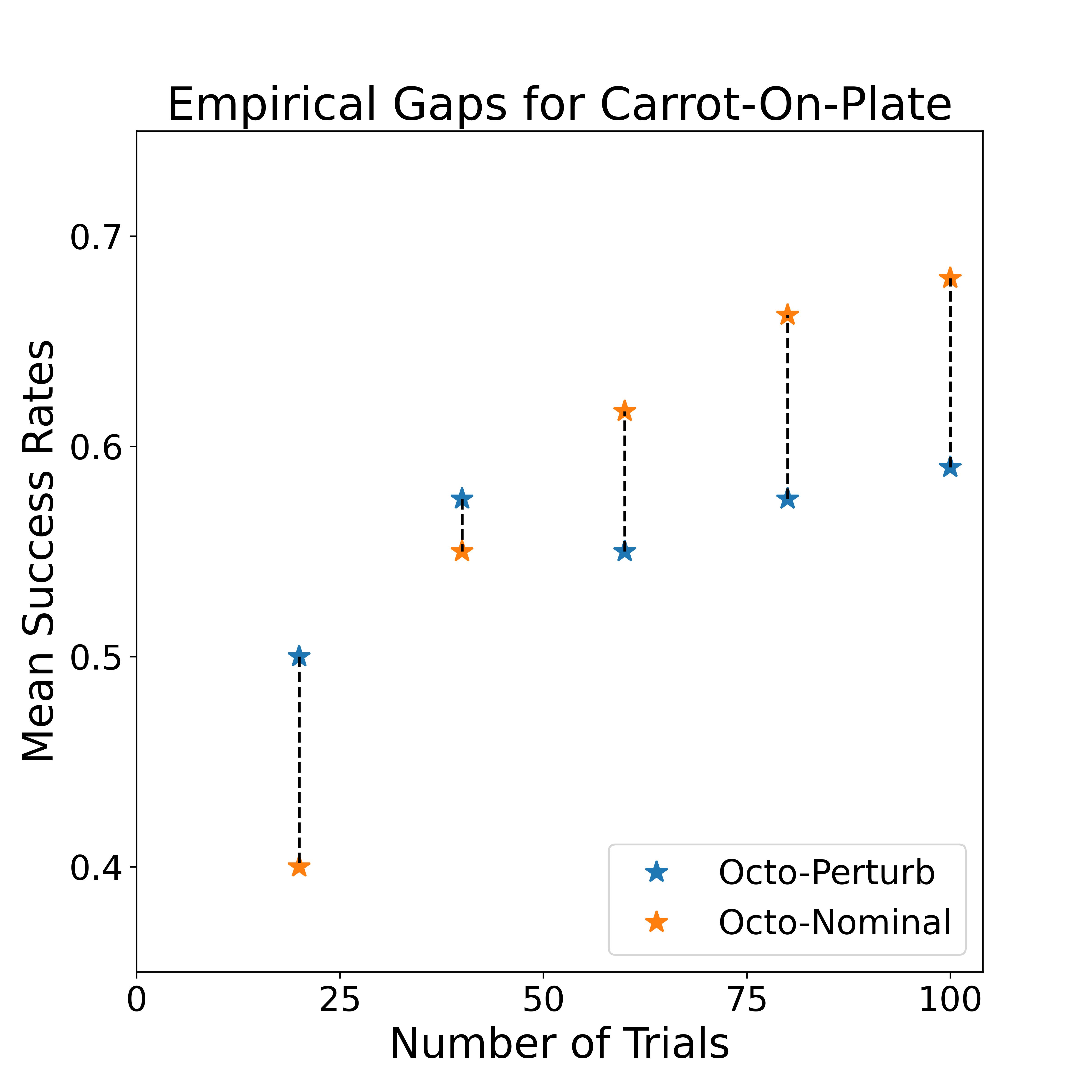}
    \subcaption{Octo-Base (\textbf{Env1}) and Octo-Base (\textbf{Env2}) in real-world \textbf{CarrotOnPlate} task}
    \label{fig:CarrotPlate_Dist}
\end{minipage}
\begin{minipage}[t]{0.32\linewidth}
    \centering
    \includegraphics[width=0.9
    \textwidth]{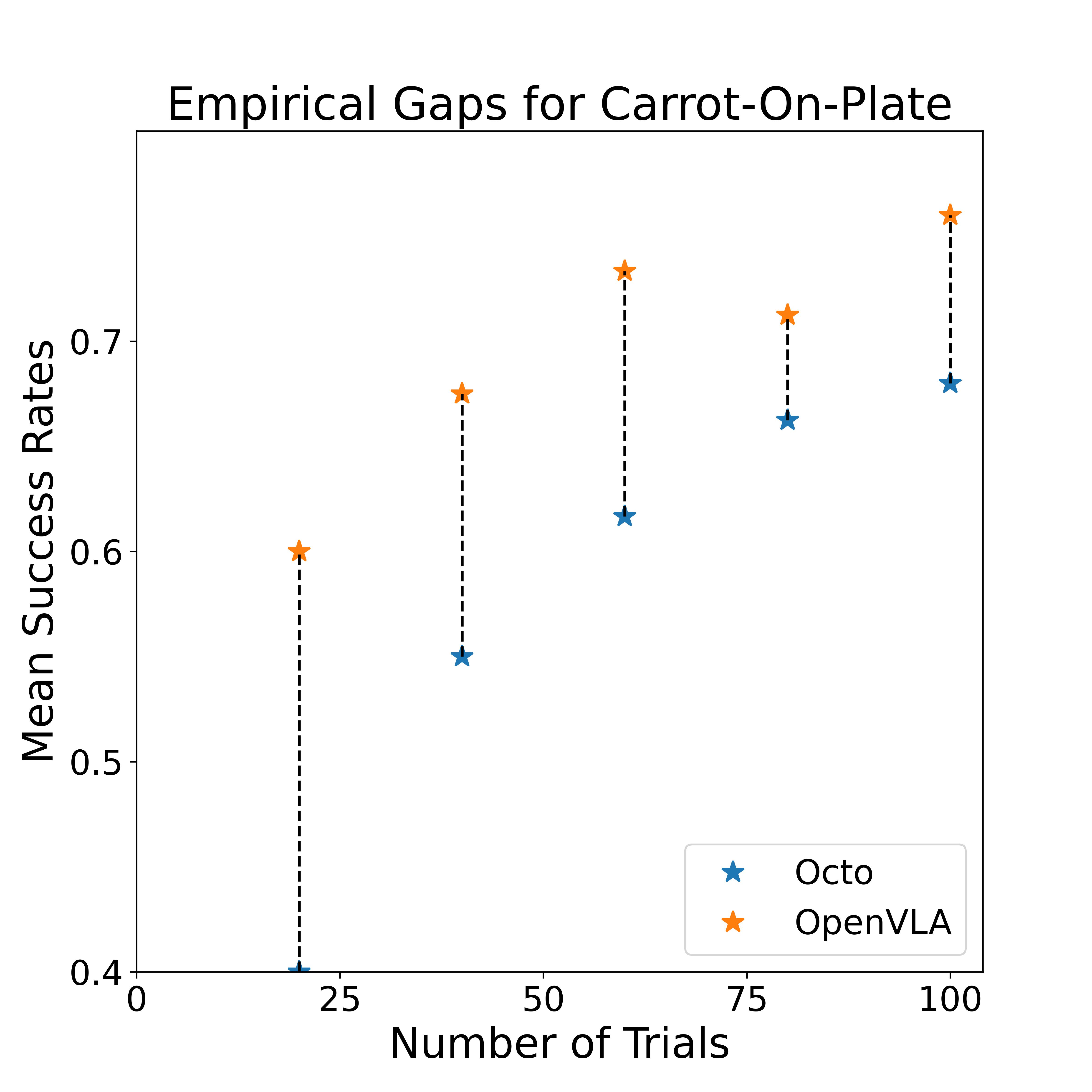}
    \subcaption{Octo-Base (\textbf{Env1}) and OpenVLA (\textbf{Env1}) in real-world \textbf{CarrotOnPlate} task}
    \label{fig:CarrotPlate_Policy}
\end{minipage}
\caption{Running empirical success rates of two policies as the number of trials increases. (a) In the \textbf{EggplantInBasket} task, there is a consistent gap in performance due to lower statistical uncertainty. This is reflected in \cref{table:all_hardware_results} (row 8) where \OurMethod~terminates at N = 119. (b and c) On the other hand, in the \textbf{CarrotOnPlate} task, the relative performance consistently fluctuates and even sometimes flips due to high statistical uncertainty arising from the close competition between two policies. This leads to even SPRT oracle requiring more than 500 trials to confidently determine the relative performance (\cref{table:carrot_on_plate}).}
\end{figure*}

We explore the results of the \textbf{CarrotOnPlate} hardware results in more detail.
\cref{fig:CarrotPlate_Dist} and \cref{fig:CarrotPlate_Policy} illustrate how the running empirical success rates change as $N$ grows. Note that the relative performance consistently fluctuates and even sometimes flips, which indicates the inherent difficulty of comparison when the two policies are closely competing. In order to estimate the minimum number of necessary trials for these challenging comparisons, we run the SPRT Oracle on multiple instances of $N_{\max}$. Namely, we assume that the true underlying distribution matches the terminal empirical success rates ($\mathbb{H}_1\text{ : }(p_0, p_1) = (0.59, 0.68)$ for \textbf{Octo} (\textbf{Env2}) vs \textbf{Octo} (\textbf{Env1}) and $\mathbb{H}_1\text{ : }(p_0, p_1) = (0.68, 0.76)$ for \textbf{Octo} (\textbf{Env1}) vs \textbf{OpenVLA} (\textbf{Env1})). We determine the worst-case point null (corresponding $\mathbb{H}_0\text{ : }(p_0, p_1) = h_0^* \in \mathcal{H}_0$) for each case and run the SPRT Oracle on the associated simple-vs-simple test, where it is essentially optimal. We observe the following empirical power results (\cref{table:carrot_on_plate}), which can be understood as approximating the probability of rejecting the null (under the draw of the sequence of i.i.d. data) at each level of $N_\text{max}$ when the true gap matches the empirical gap observed on 100 trials in hardware. 

\begin{table}[htb!]
    \centering
    \resizebox{\linewidth}{!} {
    \begin{tabular}{c|c|ccccc}
    \toprule
  Case ($\downarrow$) & $N_\text{max}$ ($\rightarrow$) & 100 & 200
 & 300 & 400 & 500 \\
     \midrule
   $(0.59, 0.68)$ & SPRT Power ($\rightarrow$) & 0.324 & 0.513 & 0.676 & 0.762 & 0.823 \\
    $(0.68, 0.76)$ &  SPRT Power ($\rightarrow$) & 0.337 & 0.491 & 0.643 & 0.724 & 0.804 \\
    \bottomrule
    \end{tabular}
    }
    \caption{Empirical power of SPRT Oracle on distributions matching the empirical gaps observed in hardware trials of \textbf{CarrotOnPlate}. This suggests that at 200 trials per policy, there is only about a 50\% chance of observing a sequence leading to rejection of the null; even for the oracle, 500 trials are required before this approximate probability reaches 80\%. }
    \label{table:carrot_on_plate}
\end{table}
As shown \cref{table:carrot_on_plate}, nearly 400 trials are required before reaching an approximately 75\% chance of rejection over the draw of observed sequences. We emphasize that this is computed via a method that is optimal with respect to the expected sample size; as such, the evaluation requirements are primarily fundamental to the variance of Bernoulli random variables, and thus represent fundamental uncertainty and sample complexity for the policy comparison problem. 

\subsection{Additional \textbf{CarrotOnPlate} Experiments}
\label{subsec:add_carrot_analysis}
A prior iteration of the \textbf{CarrotOnPlate} experiments (not reporeted in \cref{sec:experiments}) involved a hardware implementation error in which the end-effector rotation commands output by the policies were not correctly published. As a result, for the purpose of policy comparison, we label these policies \textbf{PolicyA} (in place of Octo) and \textbf{PolicyB} (in place of OpenVLA). Note that the hardware implementation does not invalidate the policy comparison procedure itself. The environment distributions \textbf{Env1} and \textbf{Env2} are as described in Section~\ref{sec:experiments}. We set \(N_{\max}=200\) as the evaluation budget for the following three settings: i) \textbf{PolicyA} (under \textbf{Env1}), ii) \textbf{PolicyB} (under \textbf{Env1}), and iii) \textbf{PolicyA} (under \textbf{Env2}). We compare \textbf{PolicyA} (\textbf{Env1}) with \textbf{PolicyB} (\textbf{Env1}), and \textbf{PolicyA} (\textbf{Env1}) with \textbf{PolicyA} (\textbf{Env2}). These results are listed in~\cref{tab:pol1vpol2}. Observe that despite utilizing the full budget, no procedure yields conclusive results in the first comparison. However, in the second comparison, we ran the evaluation after collecting 150 trials per policy. Because every method terminated, we were able to stop early at 150 trials (i.e., saving ourselves 50 hardware evaluations).
\begin{table}[t!]
    \centering
    \begin{tabular}{l|c|c|cccc}
    \toprule
    $N$ &  $\hat{p}_0$ & $\hat{p}_1$ & SAVI & Lai & STEP & SPRT$^{***}$  \\ \midrule
     200 & 0.530 & 0.560 & \text{--} & \text{--} & \text{--} & \text{--} \\
     150 & 0.613 & 0.827 & 132 & 64 & 61 & 26 \\
     \bottomrule
    \end{tabular}
    \caption{Additional \textbf{CarrotOnPlate} evaluations with $N_{\max} = 200$. \textit{Row 1:} Comparing \textbf{PolicyA} under \textbf{Env1} (\(\pi_1\)) with \textbf{PolicyA} under \textbf{Env2} (\(\pi_0\)). \textit{Row 2:} Comparing \textbf{PolicyA} under \textbf{Env1} (\(\pi_0\)) with \textbf{PolicyB} under \textbf{Env1} (\(\pi_1\)).}
    \label{tab:pol1vpol2}
    \vspace{-5mm}
\end{table}

\subsection{Mathematical and Numerical Notes}
\label{subsec:Math}
\subsubsection{Worst-Case Null Hypotheses}
The worst-case null hypotheses are computed in this framework as the real number $p \in (0,  1)$ that maximizes the expected log-likelihood ratio. First, noting the monotonicity properties of the joint distribution, we claim that the worst-case null hypothesis must lie on the line $p_0 = p_1 = p \in (0, 1)$. Second, noting the optimal power properties of the SPRT for simple-vs-simple problems, we construct the log probability-ratio test maximization as: 
\begin{equation*}
\begin{split}
    \argmax_p & \mathbb{E}_{x \sim (p, p)}\left[(\frac{p_1}{p})^x (\frac{1-p_1}{1-p})^{1-x} (\frac{p_0}{p})^x (\frac{1-p_0}{p})^{1-x}\right] \\
    \equiv \argmax_p & \mathbb{E}\left[ x \log{\frac{p_0 p_1}{p^2}} + (1-x) \log{\frac{(1-p_0)(1-p_1)}{(1-p)^2}}\right]
    \end{split}
\end{equation*}
Differentiating, the solution is the interpolation in the natural parameter space of the Bernoulli distribution: 
\begin{equation*}
\begin{split}
    \log{\frac{p^*}{1-p^*}} & = \frac{\log{\frac{p_0}{1-p_0}} + \log{\frac{p_1}{1-p_1}}}{2} \\
    \implies \eta^* & = \frac{\eta_0 + \eta_1}{2};
    \end{split}
\end{equation*}
the reconstruction of $p^*$ follows directly as
\begin{equation*}
    p^* = (1 + \exp{\eta^*})^{-1}.
\end{equation*}
That is, the worst-case null in the sense of `falsely' maximizing the probability ratio test under the null is precisely the interpolation in natural parameter space of $(p_0, p_1)$. In fact, the `true' worst-case null is difficult to compute exactly; as we verify the Type-1 Error control against the additional methods of linear projection in the nominal parameter space
\begin{equation*}
    p^* = \frac{p_0 + p_1}{2}
\end{equation*}
and as the interpolation under the KL-divergence `pseudo-distance:' 
\begin{equation*}
    p^* = \big\{p' \in (0, 1) \text{ } \rvert \text{ KL}(p_0, p') = \text{KL}(p_1, p') \big\}.
\end{equation*}
In practice, assuming continuity corrections are applied to any case in which $p_0$ or $p_1$ belong to $\{0, 1\}$, these methods generally result in similar estimates of the worst-case null hypothesis, and form a small region in which the error control can be verified to greater numerical accuracy. 

\begin{table*}[t!]
    \centering
    \resizebox{\linewidth}{!}{%
    \begin{tabular}{l|c|ccc|cc|ccc|c}
    \toprule
    Task & Type & $\alpha^*$ & $N_{\text{max}}$ & $N$ &  $\hat{p}_0$ & $\hat{p}_1$ & SAVI & Lai & STEP (Ours) & SPRT$^{***}$ \\
    \midrule
    FoldRedTowel &  $\pi_i$ & 0.05 & 50 & 50 & 0.560 & 0.920 & 21.1 (0.63) & 19.3 (0.49) & 18.9 (0.44) & 14.5 (0.53) \\
    FoldRedTowel &  $\pi_i$ & 0.05 & 200 & 50 & 0.560 & 0.920 & 21.1 (0.63) & 28.6 (0.53) & 24.4 (0.53) & 14.5 (0.53) \\
    FoldRedTowel &  $\pi_i$ & 0.05 & \underline{500} & 50 & 0.560 & 0.920 & 21.1 (0.63) & 32.0 (0.56) & 28.3 (0.56) & 14.5 (0.53) \\
    CleanUpSpill &  $\mathcal{D}^{i}_{s_0, o_0}$ & 0.05 & 50 & 50 & 0.280 & 0.800 & 13.8 (0.44) & 13.6 (0.33) & 14.0 (0.30) & 11.4 (0.41)  \\
    CleanUpSpill &  $\mathcal{D}^{i}_{s_0, o_0}$ & 0.05 & 200 & 50 & 0.280 & 0.800 & 13.8 (0.44) & 20.4 (0.38) & 18.2 (0.38) & 11.4 (0.41)  \\
    CleanUpSpill &  $\mathcal{D}^{i}_{s_0, o_0}$ & 0.05 & \underline{500} & 50 & 0.280 & 0.800 & 13.8 (0.44) & 23.6 (0.44) & 20.7 (0.43) & 11.4 (0.41) \\
    CarrotOnPlate & $\mathcal{D}^{i}_{s_0, o_0}$ & 0.05 & 100 & 100 & 0.590 & 0.680 & \text{--} & \text{--} & \text{--} & \text{--} \\
    CarrotOnPlate & $\pi_i$ & 0.05 & 100 & 100 & 0.680 & 0.760 & \text{--} & \text{--} & \text{--} & \text{--}  \\
    \midrule
    SpoonOnTowel & $\pi_i$ &  0.01 & 500 & 500 & 0.084 & 0.386 & 44.6 (1.25) & 59.8 (1.17) & 49.6 (1.10) & 34.5 (1.08) \\
    EggplantInBasket & $\pi_i$ & 0.01  & 500 & 500 & 0.400 & 0.564 &  233.2 (6.7) & 202.4 (5.0) & 183.9 (4.9) & 193.4 (5.3) \\
    StackCube & $\pi_i$ & 0.01  & 500 & 500 & 0.000 & 0.030 &  328.2 (4.8) & 409.8 (4.3) & 265.5 (4.2) & 70.5 (3.0)  \\
    \midrule
    Multitask & $\pi_i$ & 0.03 & 1500 & 1500 & \text{N/A} & \text{N/A} & 606.0 & 672.0 & 499.0 & 298.4 \\
    \bottomrule
    \end{tabular}
    }
    \caption{Empirical \textbf{expected} time-to-correct-decision for all \textbf{hardware (top)} and \textbf{simulation (bottom)} policy comparisons. The contexts, parameters, and annotation are identical to those in~\cref{table:all_hardware_results}. Summary statistics are taken over 400 random trajectories generated according to a Bernoulli distribution with data-generating (i.e., `true') parameters corresponding to the observed empirical success rates ($\hat{p}_0, \hat{p}_1$). We report the average stopping times of all methods on the right of the table for every context (standard deviation of the \emph{empirical mean} in parentheses). In all cases: lower is better. In the Multitask setting, we test $p_1 > p_0$ \textit{uniformly across the preceding three tasks}. This stopping time is the sum by column of the average stopping times for the three tasks. Our method saves the evaluator approximately 110 to 170 trials (in expectation) in uniform certification over these three tasks as compared to either feasible baseline. Note that for any single sequence of evaluations, the standard deviation for the stopping time on a given task can be approximately computed by multiplying the parenthetical standard deviation by $20$ (the square root of the number of trials (400)). This correction confirms that the observed improvement of 160 evaluation trials saved in multitask simulation in \cref{table:all_hardware_results} is consistent with the results observed here.}
    \label{table:averaged_bernoulli_counterfactual_hardware_results}
\end{table*}

\subsubsection{Discretizing the Null Hypotheses} 
\label{subsubsec:null_discretization}
In order to discretize the null hypotheses safely, it is necessary to ensure coverage over the set of possible worst-case nulls: $\{(p, p): p \in[0, 1]\}$. First, we establish an interior bound $(\epsilon, 1 - \epsilon)$ to the necessary values $p \in (0, 1)$. Specifically, for a fixed $N_\text{max}$ one can derive a value of $\epsilon$ such that if $p \geq 1 - \epsilon$ (or $p \leq \epsilon)$, it holds w.p. $\geq 1-\alpha^*$ that $\hat{p}_{1, n} = 1$ (resp. $0$) for all $n \in \{1, ..., N_\text{max}\}$. These extremal nulls pose no risk to the algorithm (because they cannot violate $\alpha^*$ Type-I error if we never \textbf{RejectNull} when $\hat{p}_1 \leq \hat{p}_0$). With this limitation in place we avoid problems arising from the rapid decay of the variance near $0$ and $1$ in the distribution set. Now, discretization in the range $(\epsilon, 1-\epsilon)$ can be undertaken to approximate all possible worst-case null hypotheses. In practice, we used approximately 100 points for $N_\text{max}$ up to $500$; this is significantly (3x) more than the default in the Scipy implementation of Barnard's Test \cite{2020SciPy-NMeth}. Note that formally, this discretization can be ensured to be safe numerically by using Pinsker's inequality to relate the total variation distance (which upper bounds, for example, the event of a false rejection from a null hypothesis) to the KL-divergence; the implication of the inequality is that the false rejection rate error due to discretization is upper bounded by a monotonic function of the maximal KL divergence between any adjacent points in the discretization; for a sufficiently dense discretization, the error can be made arbitrarily small.

\subsubsection{Technical Setting}
The problem formulation in \cref{sec:problem} was presented informally to avoid needless over-technical confusion. Slightly more precisely, we assume necessary measurability conditions on the random variables representing the success or failure of the policy in the environment. Given the probability space implicit in this assumption, the concatenation of observations constitutes the natural filtration on this space; that is, $\mathcal{F}_n = \{n, Z_1, Z_2, ..., Z_n\}$. In practice, knowledge of the sufficient statistic for exponential families induces us to use the compressed filtration $\mathcal{F}^{[comp]}_n = \{n, \sum_{i=1}^n z_{0, i}, \sum_{i=1}^n z_{1, i}\}$. Interestingly, using the Neyman-Pearson lemma, one can show that the two-dimensional state represents a lossy compression as a three-dimensional state is needed to construct the optimal exact SPRT.  

\subsubsection{Intuition for Tests}
We quickly summarize a few examples of extremal test procedures that can help provide scaffolding for the reader in terms of understanding the tradeoffs inherent between Type-I Error, Type-II Error, and expected sample size. First and foremost, safety is always possible in the Type-I sense: simply never reject the null (i.e., without looking at any data). Slightly more subtly, safety and small sample size is always feasible, as shown in \cref{footnote} in \cref{sec:problem}: decide without looking at any data, but first generate an independent random number uniformly on $[0, 1]$ and reject if the number is less than $\alpha^*$, otherwise fail to reject. Power and small sample sizes can be obtained accordingly at the cost of violating Type-I Error (just reject instead of failing to reject). Power-1 tests finish out the last leg of the triangle -- waiting an arbitrarily long time can allow for simultaneous control of Type-I and Type-II error (the N-P Lemma only concludes that in the batch setting -- where $N$ is fixed and finite -- there exist instances for which Type-I and Type-II Error cannot be simultaneously controlled). 

\end{document}